\documentclass{bmvc2k}

\usepackage{booktabs}
\usepackage{amssymb}
\usepackage{amsmath}
\usepackage{enumitem}
\usepackage{comment}

\usepackage{amsmath}
 
\DeclareMathOperator*{\argminA}{arg\,min}

\newcommand{\cam}[1]{\textcolor[rgb]{0,0,0} {#1}}
\newcommand{\fengx}[1]{\textcolor[rgb]{0,0,0} {#1}}
\newcommand{\revise}[1]{\textcolor[rgb]{0,0,0} {#1}}
\newcommand{\minor}[1]{\textcolor[rgb]{0,0,0} {#1}}

\title{MorphoStyle: Motion Style Transfer with Morphology Control}

\addauthor{Xin Feng}{xfeng4@ed.ac.uk}{}
\addauthor{Eleonora D'Arnese}{eleonora.darnese@ed.ac.uk}{}
\addauthor{Mohan Sridharan}{m.sridharan@ed.ac.uk}{}

\addinstitution{
 Institute of Perception, Action, and Behavior\\
 School of Informatics\\
 University of Edinburgh\\
 Edinburgh, UK
}

\runninghead{X. Feng, E. D'Arnese, M. Sridharan}{MorphoStyle}

\def\etal{\emph{et al}\bmvaOneDot}

\begin{document}

\maketitle
\vspace{-12pt}
\begin{abstract}
\fengx{Human motion may be viewed as a combination of action content, style, and body morphology. Existing motion style transfer methods transfer a reference style onto a content motion while assuming a canonical body, whereas shape-aware motion generators adapt motion to a target shape without explicit style control. This separation of motion style and shape (morphology) makes it difficult to generate stylized motions for non-canonical bodies; naively combining a style transfer module with a shape-aware generator often leaks action content from the style reference and disrupts shape-consistent kinematics.}
We present \textit{MorphoStyle}, a framework for shape-aware motion style transfer that is built on a shape-conditioned \fengx{Finite-Scalar-Quantization Variational Auto-Encoder (FSQ-VAE)}. The key contribution is to pose the desired style transfer as modular latent disentanglement comprising: (i) a contrastive style encoder that extracts content-decoupled style embeddings; (ii) a text-guided style-routing mechanism that locates style-relevant joints in a text-motion feature space; and (iii) a manifold preserving style modulator that injects discriminative style embeddings in content features as a temporally-gated low-rank offset. Extensive experiments on benchmark datasets demonstrate that MorphoStyle outperforms competing baselines in terms of both shape control and motion style transfer, while simultaneously providing quantitative shape control. For more details, please see project website: \href{https://github.com/funkdub/MorphoStyle}{https://github.com/MorphoStyle}.
\end{abstract}

\section{Introduction}
\label{sec:intro}

Human motion generation aims to synthesize realistic and natural human movements conditioned on various inputs, such as text descriptions, reference motion, audio, or body parameters. It is a core research topic in computer vision and graphics, and has extensive applications in gaming~\cite{holden2020learned}, film production~\cite{yamane2010animating}, virtual reality~\cite{holden2017phase}, and robotics~\cite{peng2021amp}. 
\fengx{Realistic human motion is determined by three complementary factors: the content, \emph{i.e.}, what action is performed, such as walking or sitting down; the style, \emph{i.e.}, how the action is performed, such as a funny or childish gait; and the morphology, \emph{e.g.}, who performs the action, such as a tall and lean adult or a short and heavy one. All three components are essential for realistic and personalized digital avatars across various applications, since identical content with mismatched style or morphology breaks the perception of realism. Despite rapid progress in human motion generation based on deep generative models~\cite{ho2020ddpm,tevet2023mdm,chen2023mld,guo2024momask,guo2022humanml3d,petrovich2022temos}, jointly controlling transfer of motion style and body shape remains a challenging open problem.} 

Most existing methods for motion synthesis focus on either motion style transfer or shape-aware motion generation. Motion style transfer methods aim to generate a novel motion that adapts the content obtained from a source to the style of a reference motion~\cite{holden2016deep,holden2017fast,aberman2020unpaired,xia2015realtimestyle,brand2000stylemachines,li2002motiontexture,jang2022motionpuzzle,tao2022style,guo2024generative,zhong2024smoodi,guo2025stylemotif,li2024mulsmo}.
Due to the difficulty of capturing paired stylistic motions, most recent methods start with pretrained latent-diffusion backbones~\cite{chen2023mld,tevet2023mdm} and inject style through ControlNet-style adaptors~\cite{zhong2024smoodi,zhang2023controlnet} or cross-modal alignment with foundation embeddings~\cite{guo2025stylemotif,girdhar2023imagebind}. While these methods produce visually realistic stylized motions, they have two main limitations. First, style features are fused with the content stream of a pretrained latent-diffusion space without explicit disentanglement of content and style, resulting in content leakage and unrealistic kinematics in the synthesized output~\cite{zargarbashi2026vq}. Second, the transfer of motion style is not adapted to body shape, e.g., \fengx{a canonical body is assumed and parameterized by the Skinned Multi-Person Linear (SMPL) model}~\cite{loper2015smpl}, producing unpleasant artifacts when used with non-canonical body morphologies. Meanwhile, shape-aware motion generation aims to control body shape in the motion generation pipeline~\cite{liao2025shapemymove,peng2018deepmimic,peng2021amp}. One recent example, ShapeMyMove~\cite{liao2025shapemymove}, employs a shape-conditioned Finite-Scalar-Quantization Variational Auto-Encoder (FSQ-VAE)~\cite{mentzer2024fsq} that first quantizes motions into discrete tokens and de-quantizes them in a shape-conditioned manner. This formulation does not support style control, and morphology-aware motion style transfer remains an open problem. 

We describe \textit{MorphoStyle}, a novel shape-aware motion style transfer framework that seeks to address the limitations of existing work. We introduce a content-decoupled '\textit{style encoder}' that builds on a pretrained shape-conditioned FSQ-VAE generator and injects style embeddings into motion content. It does so through a contrastive learning-based style encoder that maps reference style motion to a content-decoupled latent feature space; and a \fengx{manifold-preserving style modulator that injects the style embeddings into content features as a temporally-gated low-rank offset.}
Specifically, the contributions of this paper are:
\begin{itemize}[itemsep=-3pt, topsep=0pt]
    \item \fengx{A unified framework for motion style transfer with morphology control. MorphoStyle builds on a frozen shape-conditioned FSQ-VAE, preserving its content and shape generation capability while introducing a dedicated branch for controllable style.}

    \item \fengx{A content-decoupled style injection mechanism that leverages a supervised contrastive encoder to learn discriminative style embeddings; a text-guided routing module to localize style-relevant joints; and a manifold-preserving style modulator to inject style embeddings into content features.}

    \item \fengx{Extensive experiments on HumanML3D and 100Style (benchmark) datasets to demonstrate better motion style transfer and morphology control, and significantly higher efficiency, compared with state-of-the-art diffusion-based baselines.}
\end{itemize}
We begin with a description of related work (Section~\ref{sec:rw}), followed by a description of our framework (Section~\ref{sec:method}), experimental evaluation (Section~\ref{sec:exp}), and conclusions (Section~\ref{sec:conclusion}).

\vspace{-4pt}
\section{Related Work}
\vspace{-4pt}
\label{sec:rw}
We motivate the proposed framework by reviewing related work in motion style transfer
and shape-aware motion generation.

\vspace{-4pt}
\subsection{Motion Style Transfer}
Motion style transfer aims to adapt motion content, \emph{i.e.} the action performed, to a style reference, \emph{i.e.} how the action is performed, to obtain a stylized output~\cite{aberman2020unpaired,holden2016deep,jang2022motionpuzzle}. Early efforts were based on statistical or graph-based methods~\cite{brand2000stylemachines,li2002motiontexture,xia2015realtimestyle}, but they provide limited expressivity in terms of the styles they can reproduce. Deep network models~\cite{goodfellow2014gan,kingma2014vae,ho2020ddpm} provided better performance through their powerful ability to interpolate over the space of training samples. Inspired by image stylization, Holden \emph{et al.}~\cite{holden2016deep,holden2017fast} performed Gram-matrix optimization on a learned motion manifold, and Aberman \emph{et al.}~\cite{aberman2020unpaired} introduced a Generative Adversarial Network model with Adaptive Instance Normalization (AdaIN)~\cite{huang2017adain} that disentangles content and style without paired supervision.
Subsequent methods have pursued temporally coherent style transfer through per-body-part graph-aware feature exchange~\cite{jang2022motionpuzzle}, spatial-temporal graph generators~\cite{park2021diverse}, time-series Transformers~\cite{tao2022style}, and pre-trained auto-encoder latent spaces~\cite{guo2024generative}.
Although these methods have improved the performance of motion style transfer, they still struggle with limited expressivity and realism in stylized motions, with the underlying motion generators being the main bottleneck.

Building on diffusion-based models for human motion generation~\cite{tevet2023mdm,chen2023mld}, SMooDi~\cite{zhong2024smoodi} employs a ControlNet-style adaptor~\cite{zhang2023controlnet} and a classifier-based style guidance to augment a pretrained latent diffusion backbone. More recently, StyleMotif~\cite{guo2025stylemotif} further extends motion stylization to multimodal style conditioning aligned with relevant AI models such as ImageBind~\cite{girdhar2023imagebind} and MulSMo~\cite{li2024mulsmo}. Despite their visual realism, these diffusion-based methods share two main limitations. First, style features are fused into the content stream of a pretrained latent diffusion space {without explicit content-style disentanglement}, which inevitably results in content leakage and unrealistic kinematics in the stylized output. Second, they address motion style transfer in isolation from body geometry, implicitly assuming a canonical body shape, thus leading to undesirable artifacts with non-canonical morphologies. In contrast, MorphoStyle injects style embeddings obtained from a latent disentangled feature space in a frozen shape-conditioned backbone to effectively achieve high-quality motion style transfer with reliable morphological control.

\vspace{-8pt}
\subsection{Shape-aware Motion Generation}
Body shape is a critical yet often overlooked factor in human motion generation; individuals with different heights, limb proportions, and body morphology perform the same action with visibly different kinematics~\cite{liao2025shapemymove,loper2015smpl}. Most human motion generators are trained on motions of a canonical SMPL body~\cite{loper2015smpl} and thus generate motion in a shape-agnostic  manner~\cite{tevet2023mdm,chen2023mld,guo2022humanml3d,guo2024momask,petrovich2022temos}.
This decreases the natural correlation between morphology and motion, as well as propagating artifacts to downstream tasks~\cite{aberman2020skeleton,villegas2018neural,liao2025shapemymove}. Some works have begun to inject shape information into the motion generation pipeline. For example, Physics-based controllers~\cite{peng2018deepmimic,peng2021amp} embed shape parameters into character simulation but rely on physical engines and reinforcement learning, and perform poorly when scaled to large motion datasets. Shape-aware pose models~\cite{kocabas2021pare,choutas2022shapy} condition kinematic regressors on body parameters to mitigate self-intersection and contact errors. \fengx{ShapeMyMove~\cite{liao2025shapemymove}, designed for shape-aware text-to-motion generation, is most related to our work. It trains a shape-aware Finite-Scalar-Quantization model (FSQ-VAE)~\cite{mentzer2024fsq} that quantizes shape-canonical motions into discrete tokens, and de-quantizes them in a shape-conditioned manner, thereby reconstructing motions adapted to body morphology. Based on this shape-aware tokenizer, it employs a language model to jointly predict shape parameters and motion token indices to yield more realistic motions with plausible body shapes.}

Although the aforementioned methods are content- and shape-driven, it is challenging to control the style of output motion since a naive combination of a style transfer mechanism and a shape-aware backbone interferes with the simultaneous modeling of style and shape. \fengx{Our MorphoStyle model extends the shape-aware FSQ-VAE backbone of ShapeMyMove~\cite{liao2025shapemymove}, but targets a fundamentally different objective of motion style transfer with morphology control}. Instead of co-training the decoder to reconstruct stylistic motions, we freeze it and extract style embeddings in a discriminative latent feature space to achieve motion style transfer. This design choice enables MorphoStyle to inherit ShapeMyMove's shape control ability, while also supporting motion style transfer. 

\vspace{-4pt}
\section{Method}
\vspace{-4pt}
\label{sec:method}

\begin{figure*}[!t]
    \centering
    \includegraphics[width=1\textwidth]{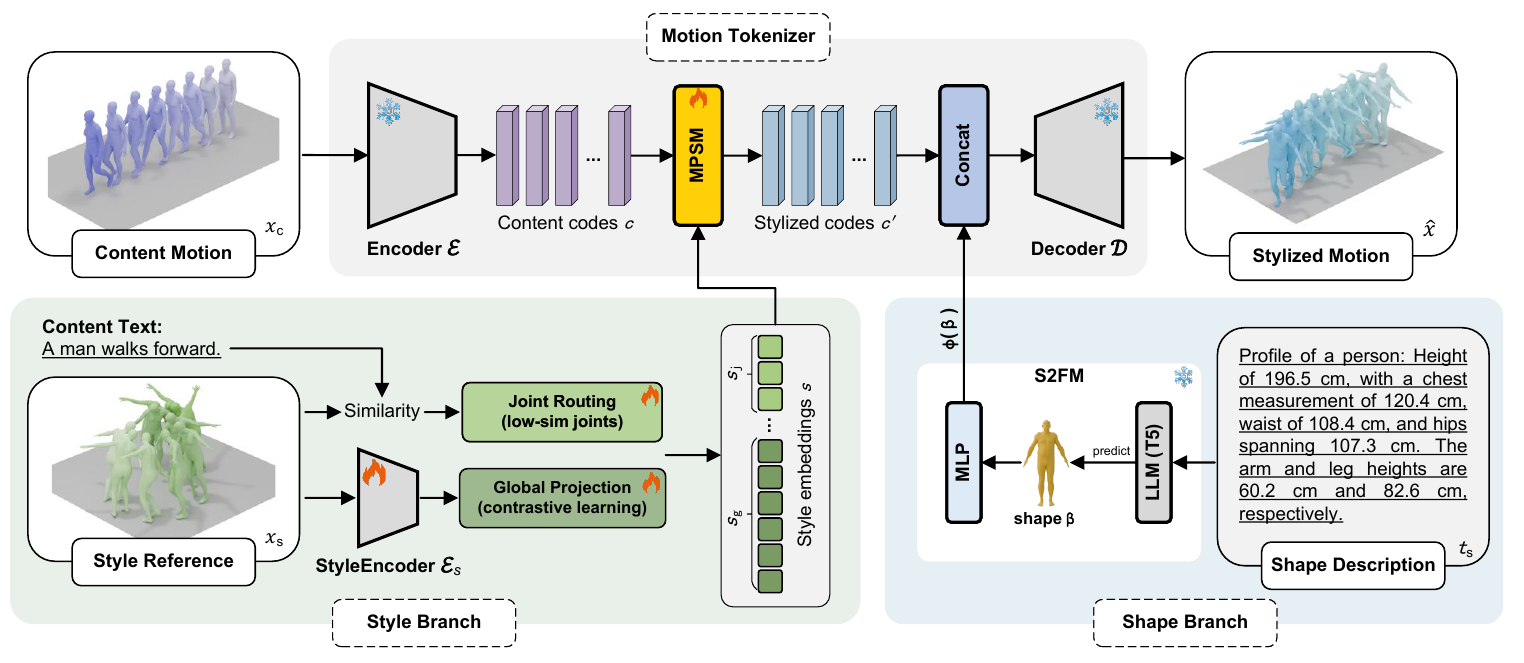}
    \caption{\revise{\textbf{Framework Overview.} Given a content motion $x_c$, a style reference $x_s$, and a textual body-shape description $t_s$, \textbf{MorphoStyle} synthesizes an output motion $\hat{x}$ that preserves the action of $x_c$, transfers the style of $x_s$, and matches the target body shape. A \emph{frozen} shape-aware FSQ-VAE tokenizes $x_c$ into content codes $c$ and decodes $\hat{x}$ via a shape-conditioned decoder $\mathcal{D}$. A \emph{trainable} style branch produces a global style embedding $s_g$ and a joint-localized embedding $s_j$, fuses them into style code $s$, and injects $s$ in $c$ through a Manifold-Preserving Style Modulator (MPSM) as a temporally-gated low-rank residual. A \emph{frozen} Shape-to-Feature module (S2FM) maps $t_s$ to shape embedding $\phi(\beta)$ conditioning $\mathcal{D}$.}}
    \label{fig:framework}
    \vspace{-12pt}
\end{figure*}

This section describes the problem formulation and basic components of MorphoStyle (Section~\ref{sec:method-framework}), and the key style disentanglement component and learning scheme (Section~\ref{sec:method-style}).

\vspace{-1em}
\subsection{Framework Overview}
\label{sec:method-framework}
\noindent\textbf{Problem formulation.}
Given content motion $x_c \in \mathbb{R}^{T\times C}$ that specifies the target action, reference motion $x_s \in \mathbb{R}^{T\times C}$ that contains the target style, and a body shape $\beta \in \mathbb{R}^{10}$ parameterized by the Skinned Multi-Person Linear (SMPL) model~\cite{loper2015smpl}, our goal is to synthesize an output motion $\hat{x} \in \mathbb{R}^{T\times C}$ that simultaneously satisfies three conditions:
(i) the action content $\hat{x}$ matches $x_c$;
(ii) the stylistic appearance of $\hat{x}$ matches that of $x_s$;
(iii) the body morphology of $\hat{x}$ is faithful to $\beta$.
\fengx{Since we adopt the standard HumanML3D feature representations~\cite{guo2022humanml3d}, per-frame feature dimension $C=263$ and sequence length $T=196$.}

\noindent\textbf{Latent disentanglement.}
MorphoStyle is built on the observation that different aspects of motion act on disjoint subspaces of a well-trained shape-aware motion auto-encoder. Specifically, in a shape-aware FSQ-VAE~\cite{mentzer2024fsq,liao2025shapemymove}, content motion is encoded in a discrete code sequence $c \in \mathbb{R}^{T' \times D}$, \fengx{where $T'=T/4$ is the temporally down-sampled length produced by the encoder's 4$\times$ temporal compression}, and code dimension $D=512$. Body-shape $\beta$ modulates de-quantization rather than the code sequence itself, and stylistic variation can be defined as a low-rank residual offset to $c$ that does not change which codebook entries are selected. Latent disentanglement indicates that \fengx{shape-aware motion style transfer can be captured by three independent modular components without shared parameters: (i) a frozen shape-aware backbone that is able to handle both content and shape; (ii) a lightweight content-decoupled style branch which stylizes the content code sequence in a low-rank manner; and (iii) a shape-to-feature module which projects textual shape description into latent shape features.}

\noindent\textbf{Our architecture.}
As presented in Figure~\ref{fig:framework}, MorphoStyle consists of a frozen shape-aware FSQ-VAE backbone, a style disentanglement branch, and a shape control branch. 
\fengx{The FSQ-VAE consisting of a content motion encoder $\mathcal{E}$, a Finite-Scalar-Quantizer, and a shape-conditioned decoder $\mathcal{D}$, is kept frozen throughout training. 
A contrastive style encoder $\mathcal{E}_s$ maps the reference style motion $x_s$ to style embeddings $s \in \mathbb{R}^{256}$, while Manifold-Preserving Style Modulator (MPSM) fuses $s$ into encoded content tokens $c$ to yield stylized motion codes $c'$.} The frozen decoder $\mathcal{D}$ then reconstructs $c'$ to the motion space conditioned on body shape $\beta$. The shape control branch uses a pretrained Shape-to-Feature module (S2FM) inspired by ShapeMyMove~\cite{liao2025shapemymove}, with a T5-based language model predicting SMPL model's $\beta$ parameters from shape descriptions $t_s$ and a Multi-layered Perceptron (MLP) model to extract shape features $\phi(\beta)=\text{S2FM}(t_s)$ from $\beta$. \minor{MorphoStyle's overall pipeline is thus}:
\vspace{-4pt}
\begin{equation}
    \label{eq:equ1}
\hat{x} \;=\; \mathcal{D}\!\left(\,\mathrm{MPSM}\!\left(\mathcal{E}(x_c),\,s\right),\,\phi(\beta)\right)
\vspace{-4pt}
\end{equation}

\noindent\textbf{Shape-aware FSQ-VAE backbone.}
MorphoStyle substantially extends the shape-aware FSQ-VAE introduced in ShapeMyMove~\cite{liao2025shapemymove}, which is pretrained on HumanML3D~\cite{guo2022humanml3d} jointly with continuous SMPL~\cite{loper2015smpl} condition parameters, and serves as a foundation of our model. Specifically, the encoder $\mathcal{E}$ employs three Conv1d residual blocks with dilation, downsamples temporal length by a factor of four, and then feeds encoded motion features into an FSQ bottleneck with a 1000-entry codebook. The shape-aware decoder $\mathcal{D}$ concatenates each quantized content code $c \in \mathbb{R}^{T'\times D}$ with a 32-dimensional shape feature $\phi(\beta)$ obtained from the shape branch, and reconstructs motions through symmetric Conv1d upsampling. Importantly, such a backbone is able to effectively control two of three factors, content and shape, thus we further introduce and optimize a lightweight style branch with Manifold-Preserving Style Modulator to inject style information without content drift.

\noindent\textbf{Shape-to-Feature Module.}
While FSQ-VAE takes $\phi(\beta)$ as its shape condition, the SMPL model's $\beta$ parameters are inconvenient since body shape is usually described in natural language. \minor{Similar to ShapeMyMove~\cite{liao2025shapemymove}}, we introduce the Shape-to-Feature Module (S2FM) (see Figure~\ref{fig:framework}) to convert textual shape description $t_s$ into shape embeddings $\phi(\beta)$:
\vspace{-4pt}
\begin{equation}
    \phi(\beta) \;=\; \mathrm{MLP}_{\phi}\!\left(\mathrm{T5}(t_s)\right)
    \vspace{-4pt}
\end{equation}
where the T5-based language model~\cite{raffel2020t5} obtains the 10-dimensional SMPL model's $\beta$ parameters from $t_s$ like ShapeMyMove~\cite{liao2025shapemymove}, and a two-layer MLP then projects $\beta$ into a 32-dimensional shape embedding that matches the shape-aware decoder $\mathcal{D}$'s input space.

\vspace{-8pt}
\subsection{Style Disentanglement}
\label{sec:method-style}
\noindent\textbf{Contrastive global style embedding.}
Contrastive style encoder $\mathcal{E}_s$ aims to extract compact and content-decoupled style embeddings that summarize whole-body stylistic cues from a reference motion. Specifically, $\mathcal{E}_s$ takes a style reference motion $x_s \in \mathbb{R}^{T\times C}$ as input. 
It consists of stacked Conv1d layers followed by adaptive average pooling, and a two-layer MLP, projecting pooled features into 256-dimensional style embeddings $s_g$.
In addition, we also introduce style dropout with probability $p=0.15$, which replaces style embeddings $s$ with a null token to improve model's robustness.
To prevent $s$ from interfering with content features, which is the root cause of the content leakage observed in previous diffusion-based stylizers~\cite{zhong2024smoodi,guo2025stylemotif}, we introduce a supervised contrastive learning mechanism that pulls together embeddings of motions with the same style label, and pushes apart embeddings of different styles in the latent feature space. \minor{The supervised contrastive learning loss~\cite{khosla2020supervised} is:}
\vspace{-4pt}
\begin{equation}
\label{equ:cl}
    \mathcal{L}_{\mathrm{cl}} = - \mathbb{E}_i \log \frac{\exp\!\left(\mathrm{sim}(s_{i},\, s_{i}^{+})/\tau\right)}{\exp\!\left(\mathrm{sim}(s_{i},\, s_{i}^{+})/\tau\right) + \sum_{s_{n}\in\mathcal{N}_i} \exp\!\left(\mathrm{sim}(s_{i},\, s_{n})/\tau\right)}
    \vspace{-4pt}
\end{equation}
where $s_{i}^{+}$ is a positive sample with the same style label as $s_{i}$; $\mathcal{N}_i$ is the set of negative samples, motions whose style labels differ from that of $s_{i}$ in the mini-batch; $\mathrm{sim}(\cdot, \cdot)$ denotes cosine similarity; and $\tau$ is a temperature hyperparameter. Contrastive supervision leads to effective disentanglement between content and style for whole-body motions in the output of $\mathcal{E}_s$.

\noindent\textbf{Text-guided joint style routing.}
Although style embeddings $s$ extract whole-body stylistic cues, they inevitably blend signals from joints that are mainly content-driven (\emph{e.g.} the pelvis during walking) with joints that are mainly style-driven (\emph{e.g.} the hands during a swaggering walk).
To explicitly disentangle them within the body, our text-guided routing module leverages a pretrained text-motion alignment model~\cite{guo2022humanml3d}, including a text encoder $\mathcal{T}$ and a motion encoder $\mathcal{M}$ jointly trained to align text and motion in a shared latent feature space. For each joint $j$ of $J=22$ body joints, we construct a joint-masked variant $\tilde{x}_s^j$ of style motion that only retains $j$'s positional, rotational, and velocity channels. We then encode each masked motion using $\mathcal{M}$ and content text caption $c_t$ using $\mathcal{T}$, and score the alignment of each joint with the caption through cosine similarity. The style-relevant joints are identified as the $K$ (=8, set experimentally) lowest values of the alignment score:
\vspace{-4pt}
\begin{equation}
    \mathcal{S} = \underset{j\in [0, J-1]}{\argminA K} \left\{ \cos\!\left(\mathcal{M}(\tilde{x}_s^{j}),\, \mathcal{T}(c_t)\right) \right\}
    \vspace{-4pt}
\end{equation}    

Given the selected set $\mathcal{S}$ of style-relevant joints, we \minor{introduce} per-joint kinematic features $\psi_j(x_s) \in \mathbb{R}^{T\times 12}$, \revise{where the $12$ channels concatenate position, rotation and velocity channels defined by HumanML3D}~\cite{guo2022humanml3d}, through an MLP $f_{\mathrm{j}}$ with an additive joint-identity embedding:
\vspace{-4pt}
\begin{equation}
\label{equ:sj}
    s_j \;=\; f_{\mathrm{j}}\!\left(\frac{1}{|\mathcal{S}|}\sum_{j\in\mathcal{S}}\big(\bar{\psi}_j(x_s) + u_j\big)\right)
    \vspace{-4pt}
\end{equation}
where $\bar{\psi}_j(x_s) = \tfrac{1}{T}\sum_{t}\psi_j(x_s)[t,:]$ are time-averaged features and $u_j$ is a learnable identity embedding that \revise{lets $f_{\mathrm{j}}$ tell joint-specific stylistic patterns apart}. Finally, we fuse joint-level embeddings with global embeddings through a residual MLP to produce unified codes $s \in \mathbb{R}^{256}$ of style reference for MPSM's style modulation:
\vspace{-4pt}
\begin{equation}
\label{equ:e6}
        s \;=\; s_g + \alpha \cdot \mathrm{MLP}\!\left([\,s_g\,;\, s_j\,]\right)
        \vspace{-4pt}
\end{equation}
where $[\cdot\,;\,\cdot]$ represents channel-wise concatenation and $\alpha=1$ is a residual scale.
At inference, the final MLP layer is initialized with a small Gaussian standard deviation of $0.01$ so that the routing residual is near zero at the resumed step, and $\alpha$ becomes a runtime knob that smoothly trades stylization strength against motion fidelity without retraining.

Intuitively, $\mathcal{M}(\tilde{x}_s^{j})$ encodes isolated motion dynamics of joint $j$, and $\mathcal{T}(c_t)$ summarizes target action from content textual caption. 
\revise{Joints whose isolated motion is least aligned with the content caption carry stylistic information that the caption does not describe, and are therefore the most informative for style transfer.} Based on the selected set of stylistic joints $\mathcal{S}$, we project kinematic features~\cite{guo2022humanml3d} $\psi_j(x_s)$ through the layers of an MLP, and then fuse it with global style embeddings $s_g$ to obtain stylistic motion features that contain the style cues of movements of style-bearing joints, while also mitigating content leakage.

\begin{figure*}[!t]
    \centering
    \includegraphics[width=0.75\textwidth]{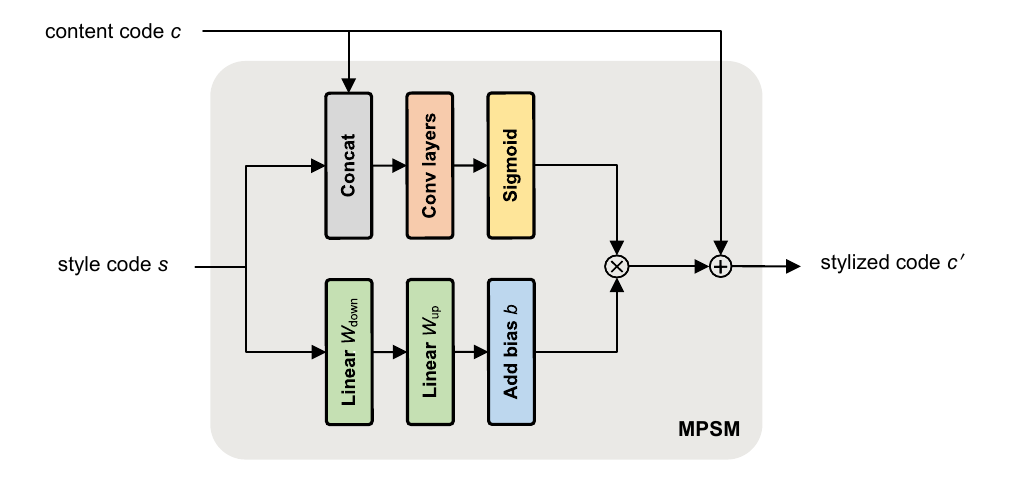}
    \vspace{-8pt}
    \caption{\fengx{Structure of Manifold-Preserving Style Modulator (MPSM).}
}
    \label{fig:mpsm}
    \vspace{-12pt}
\end{figure*}
\noindent\textbf{Manifold-Preserving Style Modulator.}
Given content codes $c = \mathcal{E}(x_c) \in \mathbb{R}^{T'\times 512}$ and style embeddings $s$, as presented in Figure~\ref{fig:mpsm}, MorphoStyle incorporates a Manifold-Preserving Style Modulator (MPSM) to fuse them into stylized codes $c'$ that remain in the codebook neighborhood of the frozen decoder throughout the motion sequence. We design the MPSM with two main requirements:
(i) style injection should be constrained to a low-dimensional subspace of each 512-dimensional code space so as not to change content information;
(ii) per-frame intensity of style injection should adapt to local kinematic context so as not to corrupt physical constraints such as foot contacts.
MPSM is thus composed of two main components: a low-rank additive offset and a learnable temporal gate.

Instead of using affine modulation schemes such as AdaIN~\cite{huang2017adain}, which multiply content features by style-conditioned scales, we project the style offset into a rank-$r$ subspace using a bilinear bottleneck in our framework:
\vspace{-4pt}
\begin{equation}
    r(s) = W_{\mathrm{up}}\, W_{\mathrm{down}}\, s + b
    \vspace{-4pt}
\end{equation}
where $W_{\mathrm{down}} \in \mathbb{R}^{r \times 256}$ and $W_{\mathrm{up}} \in \mathbb{R}^{512 \times r}$ are transformation matrices, $b \in \mathbb{R}^{512}$ is a learnable bias initialized to zero, and rank is empirically set to $r=16$.
Subsequently, we introduce a learnable temporal gate $g \in (0,1)$ calculated from both content codes $c$ and fused style embeddings $s$ to obtain per-frame intensity of style injection:
\vspace{-4pt}
\begin{equation}
\label{equ:tem}
     {g} = \sigma\!\bigl(f_{\mathrm{g}}\bigl([\,c\,;\, {s}\,]\bigr)\bigr)
     \vspace{-4pt}
\end{equation}
where $\sigma(\cdot)$ denotes the sigmoid function, $[\,\cdot~;\,\cdot]$ denotes channel-wise concatenation, and $f_{\mathrm{g}}$ is a three-layer Conv1d block with Sigmoid Linear Unit (SiLU) activations. Stylized motion codes $c' \in \mathbb{R}^{T'\times 512}$ are obtained by adding a gated low-rank offset to content motion codes:
\vspace{-4pt}
\begin{equation}
\label{equ:lr}
    c'_t = c_t + g_t \cdot r(s),\qquad t = 1, \dots, T'
    \vspace{-4pt}
\end{equation}

\subsection{Learning Scheme}
MorphoStyle is trained using a loss function that combines content-style disentanglement ($\mathcal{L}_{\mathrm{dis}}$) and shape control  ($\mathcal{L}_{\mathrm{shape}}$). All gradients are restricted to the trainable style branch; FSQ-VAE backbone and text-motion alignment model remain frozen. All loss weights (below) are fixed and specified in Section~\ref{sec:exp-settings}.

\noindent\textbf{Content-style disentanglement} loss $\mathcal{L}_{\mathrm{dis}}$ seeks to inject style into content features without content leakage. It consists of three terms. The \emph{reconstruction term} enforces fidelity of motion and code compared with the ground truth target $x^*$, with $c^*=\mathcal{E}(x^*)$, resulting in the stylized motion output being faithful to ground truth in motion-and latent code-space.
\vspace{-4pt}
\begin{equation}
    \mathcal{L}_{\mathrm{rec}}=\|\hat{x}-x^*\|_1+\lambda_{\mathrm{code}}\|c'-c^*\|_1
    \vspace{-4pt}
\end{equation}
Code-level supervision keeps modulated codes $c'$ within FSQ-VAE's codebook neighborhood, preventing style branch from causing drastic latent shifts.  

The \emph{style embedding term} aims to learn a discriminative feature space for reference style motion via supervised contrastive learning---$\mathcal{L}_{\mathrm{cl}}$ in Equation~\ref{equ:cl}---together with a content classifier loss $\mathcal{L}_{\mathrm{cc}}$. In the latent feature space, $\mathcal{L}_{\mathrm{cl}}$ pulls motion embeddings with the same style label together and pushes apart those with different styles, so encoded style embeddings $s_g$ are without unnecessary content from the style reference. \fengx{The content classifier loss $\mathcal{L}_{\mathrm{cc}}$ prevents content leakage by employing a content classifier $f_{\mathrm{ac}}$ on stylized motion $\hat{x}$, ensuring that it belongs to the same content label $y_{\mathrm{act}}$ as the corresponding content motion.
\vspace{-4pt}
\begin{equation}
    \label{eq:cc}
    \mathcal{L}_{\mathrm{cc}} \;=\; \mathrm{CE}\!\left(f_{\mathrm{ac}}(\hat{x}),\, y_{\mathrm{act}}\right)
    \vspace{-4pt}
\end{equation}
where $\mathrm{CE}(\cdot,\cdot)$ denotes the cross-entropy. 
The style embedding term is then defined as:}
\vspace{-4pt}
\begin{equation}
    \mathcal{L}_{\mathrm{emb}} = \lambda_{\mathrm{cl}}\mathcal{L}_{\mathrm{cl}}+\lambda_{\mathrm{cc}}\mathcal{L}_{\mathrm{cc}}
    \vspace{-4pt}
\end{equation}

The \emph{style classifier guidance term} \minor{uses the $47$-class style classifier of SMooDi~\cite{zhong2024smoodi} $f_{\mathrm{sm}}$}, which is trained over the joint label space of HumanML3D~\cite{guo2022humanml3d} and 100Style~\cite{mason2022real}, to provide direct supervision on the synthesized motion. 
Without explicit alignment to this classifier, even perceptually correct stylized motions may receive low style recognition accuracy due to subtle distributional shifts. We therefore introduce a training-time cross-entropy guidance that aligns $\hat{x}$ with the target style label $y_{\mathrm{sty}}$ through $f_{\mathrm{sm}}$:
\vspace{-4pt}
\begin{equation}
     \mathcal{L}_{\mathrm{cls}} = \lambda_{\mathrm{cls}}\mathrm{CE} \!\left(f_{\mathrm{sm}}(\hat{x}),\, y_{\mathrm{sty}}\right)
     \vspace{-4pt}
\end{equation}
As a result, the disentanglement loss $\mathcal{L}_{\mathrm{dis}}$ is given by:
\vspace{-4pt}
\begin{equation}    \mathcal{L}_{\mathrm{dis}}=\mathcal{L}_{\mathrm{rec}}+\mathcal{L}_\mathrm{emb}+\mathcal{L}_{\mathrm{cls}}
\vspace{-4pt}
\end{equation}

\noindent\textbf{Shape control objective.}
Although the shape-conditioned decoder $\mathcal{D}$ is frozen, style injection can still interfere with shape signal in the final output. To enforce shape consistency between synthesized motions and shape descriptions, we leverage the parameter $\beta$ of shape descriptions and introduce three shape supervision terms.

The \emph{bone-length matching term} builds on ShapeMyMove~\cite{liao2025shapemymove} to minimize Mean-Squared Error (MSE) between bone length (BL) vectors of synthesized motion $\hat{x}$ and target shape $x^*$:
\vspace{-4pt}
\begin{equation}
    \mathcal{L}_{\mathrm{bone}} \;=\; \mathrm{MSE}\!\left(\mathrm{BL}(\hat{x}),\, \mathrm{BL}(x^*)\right)
    \vspace{-4pt}
\end{equation}

The \emph{direction-aware term} ensures that bone length changes occur along realistic directions, \emph{e.g., generating longer legs and shorter arms than canonical shapes}. It seeks to avoid synthesizing motions with bones shifting in the opposite direction to prescribed ones.
\vspace{-4pt}
\begin{equation}
\mathcal{L}_{\mathrm{dir}} \;=\; 1 - \cos\!\left(\hat{x}-x^{\beta_t},\, x^{\beta_t}-x^{\beta_n}\right)
\vspace{-4pt}
\end{equation}
where $x^{\beta_t}$ and $x^{\beta_n}$ are motions with target shape $\beta_t$ and normalized shape $\beta_n$.

The \emph{magnitude-bounding term} seeks to prevent shape shift from overshooting the ground truth shift beyond a tolerance margin $\gamma$:
\vspace{-4pt}
\begin{equation}
\mathcal{L}_{\mathrm{mag}} \;=\; \mathrm{ReLU}\!\left(\|\hat{x}-x^{\beta_n}\| - \gamma\,\|x^{\beta_t}-x^{\beta_n}\|\right)^{2}
\vspace{-4pt}
\end{equation}
where $\gamma=1.5$ (set empirically) to allow limited stylistic exaggeration beyond the ground truth shape drift. The shape control loss function is the weighted sum of three components:
\vspace{-4pt}
\begin{equation}
\mathcal{L}_{\mathrm{shape}} \;=\; \mathcal{L}_{\mathrm{bone}} + \lambda_{\mathrm{dir}}\,\mathcal{L}_{\mathrm{dir}} + \lambda_{\mathrm{mag}}\,\mathcal{L}_{\mathrm{mag}}
\vspace{-4pt}
\end{equation}
These terms propagate gradients through MPSM instead of shape-aware decoder $\mathcal{D}$, ensuring reliable shape control. The overall loss function of MorphoStyle is:
\vspace{-4pt}
\begin{equation}    \mathcal{L}_{\mathrm{total}}=\mathcal{L}_{\mathrm{dis}}+\mathcal{L}_{\mathrm{shape}}
\vspace{-8pt}
\end{equation}

\vspace{-4pt}
\section{Experiments}
\label{sec:exp}
This section describes the experimental setup and results.

\subsection{Experimental Settings}
\label{sec:exp-settings}
\noindent\minor{\textbf{Hypothesis.}
Our experiments evaluate three main hypotheses about MorphoStyle:
(\textbf{H1}) our style disentanglement mechanism transfers style more faithfully and with better physical plausibility than competing baselines;
(\textbf{H2}) style injection can be decoupled from the shape-control pathway to preserve quantitative shape control under style transfer; and 
(\textbf{H3}) MorphoStyle achieves much lower computational cost than diffusion-based stylizers.}

\noindent\textbf{Datasets.}
We conduct extensive experiments on \emph{HumanML3D}~\cite{guo2022humanml3d} for content motion and \emph{100Style}~\cite{mason2022real} for style reference, following the standard SMooDi evaluation protocol~\cite{zhong2024smoodi}. HumanML3D contains 14{,}616 motion sequences ($\sim 29$ hours) with three free-form captions per sequence, retargeted to the canonical SMPL skeleton~\cite{loper2015smpl}; we use the official train/val/test splits. 100Style contains 100 distinct stylistic walking variants performed by a single actor; for fair comparison with prior diffusion-based stylizers, we adopt the 47-class subset whose action categories overlap with HumanML3D, so that the same cross-dataset style classifier can be used for evaluation. To supervise the direction-aware shape control terms, we further use the paired neutral- and target-shape renderings of HumanML3D released by ShapeMyMove~\cite{liao2025shapemymove}, which provide $x^{\beta_n}$ and $x^{\beta_t}$ for each training motion. 

\noindent\textbf{Evaluation Measures.}
We adopt seven evaluation protocols to account for both motion style transfer quality and shape control fidelity. For motion style transfer, we follow the SMooDi protocol~\cite{zhong2024smoodi} to report three measures on $N=4209$ content-style pairs, where $4209$ content motions are the standard HumanML3D~\cite{guo2022humanml3d} test split, each randomly paired with a style reference sampled from the 100Style~\cite{mason2022real} dataset. \emph{Fr\'echet Inception Distance (FID)}~\cite{heusel2017fid} measures the distributional distance between generated and ground-truth motions in the Guo22 evaluator's feature space~\cite{guo2022humanml3d}; \emph{Style Recognition Accuracy (SRA)} measures stylization fidelity as the top-1 classification accuracy of a frozen 47-class style classifier on the generated motions; \emph{Foot Skating} measures physical plausibility as the average horizontal foot velocity in the foot-contact frames.
For shape control, \revise{we adopt the \emph{Bone-Length MAE} protocol, which measures absolute bone-length agreement with the target body, from ShapeMyMove~\cite{liao2025shapemymove} and introduce three additional measures.} \emph{Bone-Length Temporal Standard Deviation} (BoneLen-T-Std) measures intra-sequence stability of bone lengths; \emph{$\beta$-swap response} measures the bone-length shift induced by switching between two contrasting body presets (thin$\leftrightarrow$heavy and short$\leftrightarrow$tall); and \emph{$\beta$-correlation} measures the Pearson correlation between the recovered shape direction and the ground-truth shape direction. 

\noindent\textbf{Implementation Details.}
MorphoStyle is implemented in PyTorch~\cite{paszke2019pytorch} and trained on a single NVIDIA A100 GPU (80\,GB). To better fit the data distribution of the 100Style dataset, we follow ShapeMyMove~\cite{liao2025shapemymove} to fine-tune the FSQ-VAE backbone on both HumanML3D and 100Style. The text-motion alignment model and the Shape-to-Feature module are loaded from their released checkpoints~\cite{liao2025shapemymove,guo2022humanml3d} and kept frozen throughout the training procedure. We use the AdamW optimizer~\cite{loshchilov2019adamw} with $(\beta_1, \beta_2){=}(0.9, 0.99)$ and weight decay $10^{-4}$; the base learning rate is $1.5\times 10^{-5}$ with a $5\times$ multiplier for the joint-text routing modules to compensate for their near-zero initialization, a 50-step linear warmup, and cosine annealing to $10^{-6}$. We train for $120$ epochs with batch size $24$ on cross-domain content-style pairings sampled from HumanML3D and 100Style at a 1:1 ratio, and the SMooDi-classifier guidance is ramped up linearly over the first three epochs. 
\fengx{We fix all loss weights throughout training, with $\lambda_{\mathrm{code}}=0.5$, $\lambda_{\mathrm{cl}}=0.15$, $\lambda_{\mathrm{cc}}=0.05$, $\lambda_{\mathrm{cls}}=0.05$, $\lambda_{\mathrm{dir}}=0.5$, and $\lambda_{\mathrm{mag}}=0.25$.}
During inference, the residual scale is set to $\alpha{=}0.5$ in Equation~\ref{equ:e6}. 

\vspace{-8pt}
\subsection{Experimental Results}
\noindent\textbf{Motion style transfer.} We compare MorphoStyle with six representative motion style transfer baselines. \emph{MLD+Aberman et al.}~\cite{chen2023mld,aberman2020unpaired} and \emph{MLD+MotionPuzzle}~\cite{chen2023mld,jang2022motionpuzzle} are classical cascades that attach an AdaIN-based or graph-aware stylizer on top of a pre-trained text-to-motion latent diffusion model. \emph{SMooDi}~\cite{zhong2024smoodi} and \emph{StyleMotif}~\cite{guo2025stylemotif} are two recent diffusion-based stylizers that augment a latent-diffusion backbone with a ControlNet-style adaptor and cross-modal alignment (respectively). \emph{ShapeMyMove}~\cite{liao2025shapemymove} is the shape-aware text-to-motion generator on which our backbone is built. Although it has no stylization option, we compare it to isolate the contribution of the FSQ-VAE backbone alone.
\emph{SMooDi} +~\emph{ShapeMyMove} is a stronger cascade we constructed by attaching SMooDi's style adaptor to the ShapeMyMove~\cite{liao2025shapemymove} shape-aware decoder; it serves as the most direct baseline for comparison. 

As reported in Table~\ref{tab:mst}, MorphoStyle achieves the best Style Recognition Accuracy (SRA) and the best Foot Skating ratio among all competing methods, while remaining competitive on FID. Specifically, MorphoStyle attains an SRA of $81.9\%$, surpassing the baseline StyleMotif by $13.1\%$; this large gap indicates that our modular latent disentanglement is substantially more effective in injecting style into the generative latent than ControlNet-style or cross-modal-alignment alternatives. In terms of physical plausibility, our Foot Skating ratio of $0.081$ is the lowest across the board, owing to the temporal gate of MPSM that adaptively attenuates style injection at contact-critical frames. Regarding distributional realism, our FID of $2.84$ is comparable to that of SMooDi ($2.95$) and substantially lower than that of all cascade baselines ($3.89$ to $7.87$), although it does not surpass StyleMotif ($1.38$); we attribute this gap to StyleMotif's significantly heavier $\sim 495$\,M-parameter latent-diffusion backbone.
More importantly, MorphoStyle is the \emph{only} method that simultaneously delivers high-quality motion style transfer and quantitative shape control: the four diffusion-based baselines are architecturally incapable of supporting shape input, and the SMooDi+ShapeMyMove, although shape-aware in design, suffers severely on both FID and Foot Skating due to the latent-space interference. These results strongly support \textbf{H1}.

\begin{figure*}[!t]
    \centering
    \includegraphics[width=0.85\textwidth]{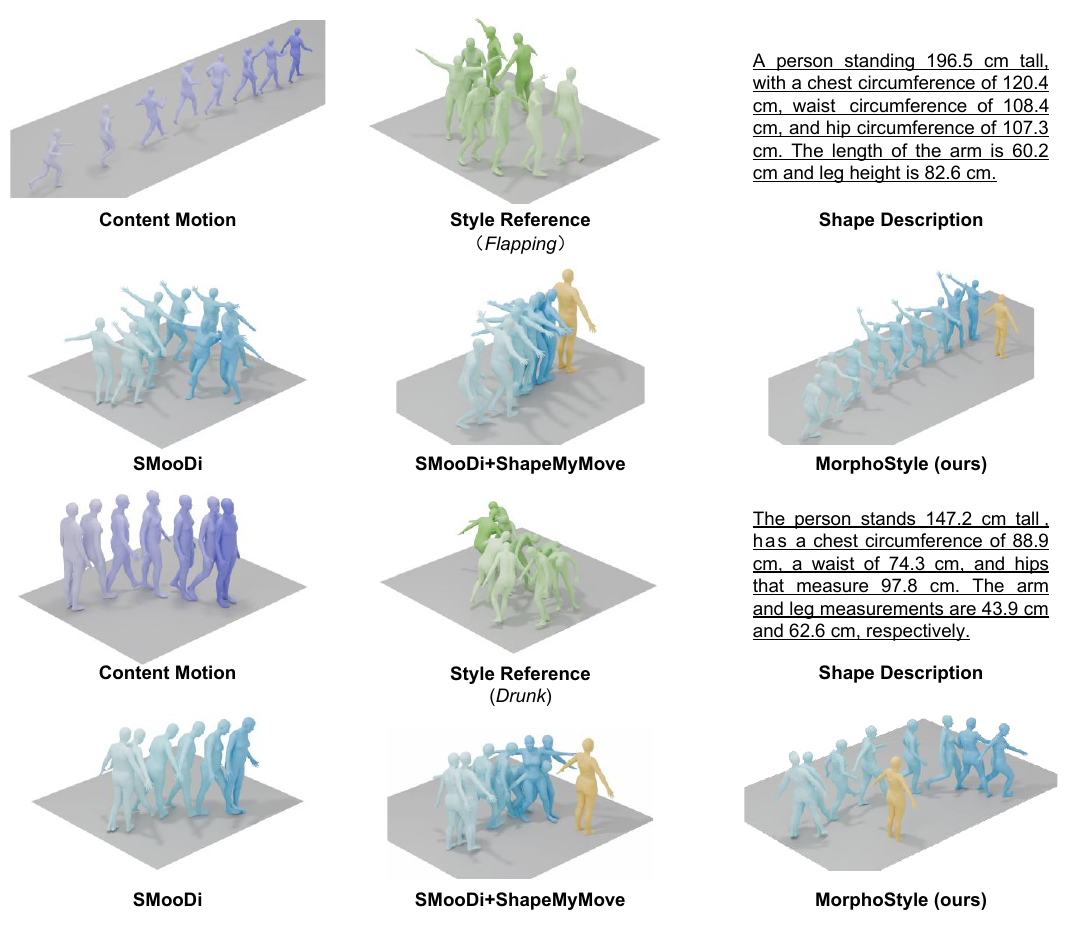}
    \vspace{-12pt}
    \caption{\cam{Qualitative comparison of shape-aware motion style transfer against SMooDi~\cite{zhong2024smoodi} and SMooDi+ShapeMyMove. Yellow body is the target shape used for reference.}}
    \label{fig:style}
    \vspace{-4pt}
\end{figure*}
\begin{table}[!t]
\centering
\small
\begin{tabular}{lcccc}
\toprule
\textbf{Method} & \textbf{SRA$\uparrow$} & \textbf{FID$\downarrow$}
                & \textbf{Foot Skate$\downarrow$}
                & \textbf{Shape Control} \\
\midrule
MLD~+~Aberman~\etal           & 61.1\% & 3.89          & 0.338          & $\times$     \\
MLD~+~Motion Puzzle           & 67.2\% & 6.87          & 0.197          & $\times$     \\
SMooDi             & 65.1\% & 2.95          & 0.095          & $\times$     \\
StyleMotif          & 68.8\% & \underline{1.38} & \underline{0.094}          & $\times$     \\
ShapeMyMove &1.57\%& \textbf{1.36} & 0.228 & $\checkmark$\\
SMooDi~+~ShapeMyMove    & \underline{69.9\%} & 7.87          & 0.245          & $\checkmark$ \\
\textbf{MorphoStyle (ours)} & \textbf{81.9\%} & 2.84 & \textbf{0.081} & $\checkmark$ \\
\bottomrule
\end{tabular}
\caption{Comparing MorphoStyle with baselines for motion style transfer on HumanML3D content with 100Style references. \textbf{Bold} indicates best result and \underline{underline} the second-best.}
\label{tab:mst}
\vspace{-8pt}
\end{table}

Figure~\ref{fig:style} provides qualitative comparisons of stylized motions conditioned by different body shape descriptions. SMooDi+ShapeMyMove transfers the coarse appearance of the reference style but corrupts the motion content, producing undesired content leakage in the form of over-extended hand swings, irregular gait timing, and noticeable foot skating. MorphoStyle reproduces the target style while remaining faithful to the input motion content throughout the sequence; the generated motion is temporally coherent and free of skating.

\noindent\textbf{Shape control.}
We evaluated shape control fidelity of MorphoStyle against two shape-aware baselines, ShapeMyMove~\cite{liao2025shapemymove} and the SMooDi+ShapeMyMove, in terms of four shape measures. We do not include SMooDi~\cite{zhong2024smoodi} or StyleMotif~\cite{guo2025stylemotif} since their backbones do not condition on body shape and are incapable of being evaluated on shape measures. As reported in Table~\ref{tab:shape}, MorphoStyle attains the lowest Bone-Length MAE ($1.91$\,cm), lowest temporal Bone-Length standard deviation ($1.26$\,cm), and highest $\beta$-swap response ($5.31$\,cm), surpassing the SMooDi+ShapeMyMove by $23.3\%$ on bone-length and a striking $61.9\%$ on temporal stability. This indicates that MorphoStyle most accurately produces motions with the target body proportions, keeps these proportions consistent over the motion sequence, and is most responsive to changes in the SMPL shape parameter $\beta$. In summary, MorphoStyle delivers the highest-fidelity shape control among all methods that simultaneously support stylization, providing strong evidence in support of \textbf{H2}.

As presented in Figure~\ref{fig:style}, we visualize motion style transfer results conditioned on different target body shapes specified by textual shape descriptions. 
SMooDi+ShapeMyMove does adapt to the target body shape, yet the bolt-on style adaptor consistently introduces drastic kinematic artifacts, confirming the architectural mismatch discussed earlier. In contrast, MorphoStyle preserves the target body morphology, produces stylistically expressive yet physically plausible motions, and remains stable across diverse style references, validating the effectiveness of our modular latent disentanglement design.
\begin{table}[!t]
\centering
\small
\begin{tabular}{lcccc}
\toprule
\textbf{Method} & \textbf{BoneLen-MAE$\downarrow$} &
                  \textbf{BoneLen-T-Std$\downarrow$} &
                  \textbf{$\beta$-swap$\uparrow$} &
                  \textbf{$\beta$-corr$\uparrow$} \\
\midrule
ShapeMyMove          & 4.59         & 3.53            & 5.05            & \textbf{0.677}             \\
SMooDi~+~ShapeMyMove & \underline{2.49}          & \underline{3.31}          & \underline{3.87}          & 0.314          \\
\textbf{MorphoStyle (ours)}        & \textbf{1.91} & \textbf{1.26} & \textbf{5.31} & \underline{0.332} \\
\bottomrule
\end{tabular}
\caption{Comparison with competing methods for shape control during motion style transfer using four evaluation measures. \textbf{Bold} indicates the best result and \underline{underline} the second-best.}
\label{tab:shape}
\vspace{-8pt}
\end{table}
\begin{table}[!t]
\centering
\small
\begin{tabular}{lccc}
\toprule
\textbf{Method} & \textbf{Trainable params}
& \textbf{Inference time$\downarrow$}
& \textbf{GFLOPS$\downarrow$} \\
\midrule
SMooDi        & 13.90 M         & 7.19s            & 79.59\\ 
SMooDi+ShapeMyMove  & 31.11 M         & 2.11s            & 112.84 \\
\textbf{MorphoStyle (ours)}      & \textbf{4.25 M} & \textbf{0.02s} & \textbf{1.98} \\
\bottomrule
\end{tabular}
\caption{Comparison of model parameters, inference speed, and computational cost.}
\label{tab:efficiency}
\vspace{-12pt}
\end{table}

\noindent\textbf{Computational overhead.}
We evaluated the computational efficiency of MorphoStyle against two competing baselines, SMooDi and the SMooDi~+~ShapeMyMove, that natively support shape control or are closest to our problem setting in terms of trainable parameter count, inference time, and computational cost measured in GFLOPs. All measurements are conducted on a single NVIDIA A100 GPU with batch size $1$ to faithfully reflect a real-time deployment scenario. \revise{Since StyleMotif~\cite{guo2025stylemotif} has not released its software or checkpoints, we do not compare it in terms of shape control (Table~\ref{tab:shape}) or computational overhead.} As reported in Table~\ref{tab:efficiency}, MorphoStyle requires $4.25$\,M trainable parameters, which is $3.3\times$ smaller than SMooDi ($13.90$\,M) and $7.3\times$ smaller than the SMooDi+ShapeMyMove ($31.11$\,M). Specifically, MorphoStyle synthesizes a stylized motion in $0.02$ second per sample (approximately $50$ FPS on a single GPU), compared to $7.19$ seconds for SMooDi's $50$-step DDIM sampler and $2.11$ seconds for the SMooDi+ShapeMyMove, yielding significant speed-up. Similarly, MorphoStyle requires only $1.98$\, GFLOPs per inference, which is significantly less than the two competing baselines. These results strongly support \textbf{H3} and are due to our design choices; unlike diffusion-based stylizers that perform iterative denoising over tens of sampling steps, MorphoStyle synthesizes the stylized motion in a single forward pass through a $4.25$\, M-parameter style branch grafted onto a frozen FSQ-VAE backbone. Such efficiency makes MorphoStyle work at real-time rates while supporting quantitative shape control, expanding its applicability to interactive animation, real-time avatar control, and on-device deployment scenarios that remain infeasible for diffusion-based competitors.

\vspace{-8pt}
\subsection{Ablation Study}
\begin{figure*}[!t]
    \centering
    \includegraphics[width=0.6\textwidth]{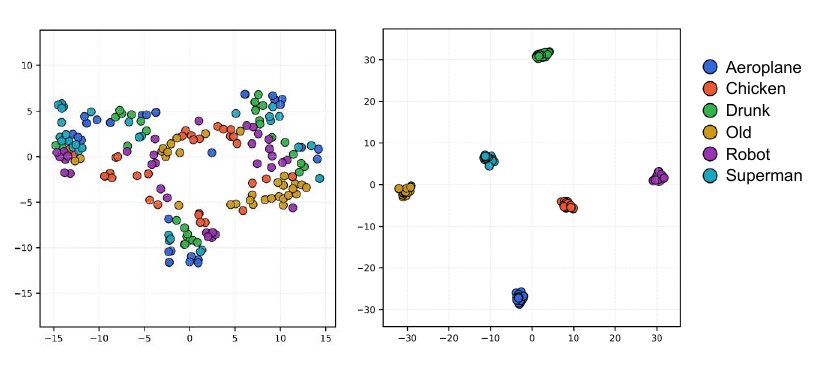}
    \vspace{-12pt}
    \caption{t-SNE visualization of global style embeddings $s_g$ on six representative style classes, \emph{with} (right) versus \emph{without} (left) the contrastive supervision $\mathcal{L}_{\mathrm{cl}}$. }
    \label{fig:tsne}
    \vspace{-4pt}
\end{figure*}

\noindent\textbf{Content-style disentanglement.}
We conducted experiments to investigate two main components of our content-style disentanglement design. The supervised contrastive supervision (\textbf{Con}) on the global style embedding $s_g$ (Equation~\ref{equ:cl}), and the text-guided joint style routing module (\textbf{JTC}) that distills $s_j$ from style-related joints (Equation~\ref{equ:sj}). Starting from a baseline variant (\textbf{Base}) that disables both, we add each component independently and finally combine them in MorphoStyle. As reported in Table~\ref{tab:cl}, adding the contrastive supervision alone yields the most improvement, and Con increases SRA by $6.5\%$, reduces FID by $0.52$, and lowers Foot Skating, which indicates that without explicit supervision in the embedding space, global style encoder $\mathcal{E}_s$ inevitably entangles content cues that degrades stylization fidelity and motion realism. Adding the joint routing brings a smaller but consistent gain, which confirms that joint-level disentanglement provides complementary local style cues that the whole-body global embedding cannot capture. Combining them in MorphoStyle provides the best performance on every measure, which demonstrates their effectiveness for high-quality motion style transfer.

\begin{table}[!t]
\centering
\small
\begin{tabular}{lccccc}
\toprule
\textbf{Variant} & \textbf{Contrastive} & \textbf{Joint}
& \textbf{FID}$\downarrow$ & \textbf{SRA}$\uparrow$
& \textbf{Foot}$\downarrow$ \\
\midrule
Base   & $\times$     & $\times$     & 3.57 & 74.7\% & 0.143 \\
JTC  & $\times$     & $\checkmark$ & 3.54 & 78.5\% & 0.139 \\
Con & $\checkmark$ & $\times$     & 3.05 & 81.2\% & 0.110 \\
MorphoStyle  & $\checkmark$ & $\checkmark$ & 2.84 & 81.9\% & 0.081 \\
\bottomrule
\end{tabular}
\caption{Ablation on content-style disentanglement components.}
\label{tab:cl}
\vspace{-12pt}
\end{table}
To further investigate the qualitative effect of contrastive supervision, we visualize global style embeddings $s_g$ with t-SNE in Figure~\ref{fig:tsne} for six representative style classes. Without contrastive supervision, embeddings of the same style class are scattered and frequently overlap with other classes, reflecting strong content entanglement in $s_g$. In contrast, with contrastive supervision, embeddings of the same style form tight and well-separated clusters in the latent feature space. This shows that our content-style disentanglement method provides a discriminative style space, which MPSM relies on for high-fidelity style injection.

\noindent\textbf{MPSM module.} 
We investigated two variants of our MPSM, low-rank additive offset (Equation~\ref{equ:lr}) and learnable temporal gate (Equation~\ref{equ:tem}). We first entirely remove MPSM (\emph{w/o MPSM}), which is replaced by a naive concatenation between content features and style embeddings, and then remove only temporal gate (\emph{w/o gate $g$}), which keeps low-rank offset but applies it uniformly across all temporal frames. As reported in Table~\ref{tab:abl}, entirely disabling the MPSM results in worse performance in terms of SRA and FID due to content leakage, which demonstrates that MPSM is an essential component for injecting style in content features. Removing the temporal gate yields a less severe but still significant degradation, indicating that applying low-rank offset across all frames corrupts key movements without temporal intensity control of style injection, degrading style fidelity and motion quality. 

\begin{figure*}[!t]
    \centering
    \includegraphics[width=0.85\textwidth]{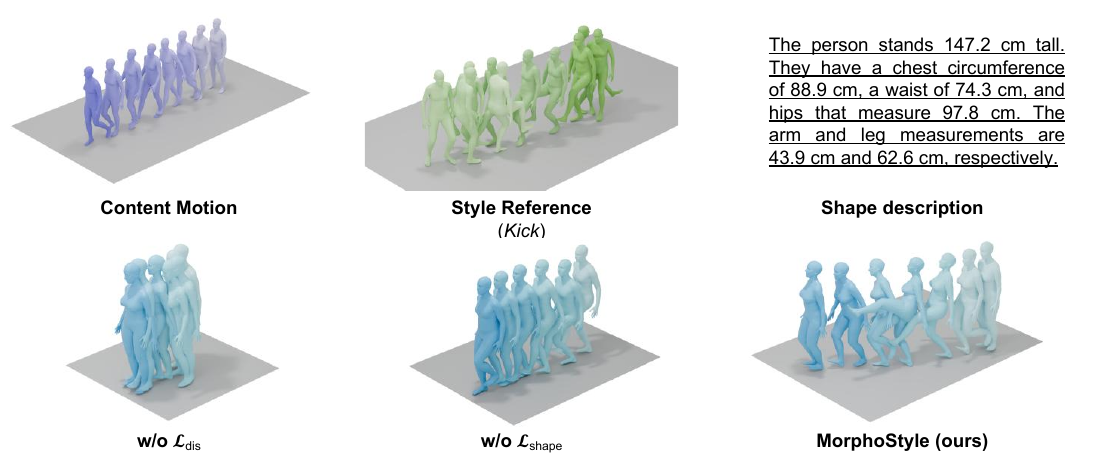}
    \vspace{-8pt}
    \caption{Qualitative ablation of two training objectives, $\mathcal{L}_{\mathrm{dis}}$ and $\mathcal{L}_{\mathrm{shape}}$, on the same content (walk forward), style reference (\emph{kick}), and target body shape.}
    \label{fig:ablation}
    \vspace{-4pt}
\end{figure*}
\begin{table}[!t]
\centering
\small
\begin{tabular}{lcccc}
\toprule
& \textbf{SRA$\uparrow$} & \textbf{FID$\downarrow$}
& \textbf{BoneLen-MAE$\downarrow$} & \textbf{$\beta$-swap$\uparrow$}\\
\midrule
w/o MPSM      &62.9\%&5.16&1.97& 2.06\\
w/o gate $g$ &72.1\%&4.85&1.92& 4.12 \\
w/o $\mathcal{L}_{shape}$  &80.4\%&2.99&2.20& 3.41\\
MorphoStyle &81.9\%&2.84&1.91& 5.31\\
\bottomrule
\end{tabular}
\caption{Ablation results on key technical components of our MorphoStyle.}
\label{tab:abl}
\vspace{-12pt}
\end{table}

\noindent\textbf{Shape control supervision.}
We also evaluated the effectiveness of the shape control objective $\mathcal{L}_{\mathrm{shape}}$ by removing it from the training process (\emph{w/o} $\mathcal{L}_{\mathrm{shape}}$) (see Table~\ref{tab:abl}). Without $\mathcal{L}_{\mathrm{shape}}$, MorphoStyle still attains a high SRA and a low FID. However, it is evident from the shape measures that Bone-Length MAE rises, and $\beta$-swap response collapses from $5.31$\ to $3.41$\, which indicates that the model no longer responds meaningfully to changes in the SMPL shape parameter $\beta$, effectively ignoring the shape signal. \revise{Figure~\ref{fig:ablation} also visualizes the failure modes of removing each loss function} confirming that the shape control objective adapts latent codes to ground truth shape direction through $\mathcal{L}_{\mathrm{shape}}$ supervision.


\vspace{-4pt}
\section{Conclusion}
\vspace{-4pt}
\label{sec:conclusion}
We presented \emph{MorphoStyle}, a unified framework for shape-aware motion style transfer that jointly models motion style transfer and body-shape control. MorphoStyle leverages a frozen shape-conditioned FSQ-VAE and injects style as a compact low-rank residual in the post-quantization content space, while preserving explicit shape control through the de-quantization pathway. This design enables effective disentanglement of content, style, and morphology with fewer parameters. Built on this backbone, our content-decoupled style branch combines contrastive style encoding, text-guided joint routing, and manifold-preserving modulation to localize and transfer style without disrupting motion content. Extensive experiments show that MorphoStyle achieves state-of-the-art performance in both motion style transfer and morphological control, while being $ 3.3\times $ smaller and $ 360\times $ faster than the closest competing baseline, enabling real-time shape-aware motion style transfer.

\smallskip
\noindent\textbf{Limitations and future work.}
MorphoStyle's distributional realism is limited by the capacity of the frozen FSQ-VAE backbone, and its FID does not surpass the diffusion-based methods. In addition, the text-guided joint routing module depends on a frozen text-motion alignment model, which may be less reliable for out-of-distribution actions. We will explore extensions to MorphoStyle to address these limitations. Future work also includes extending the framework to consider image, video, and/or audio style references, and adapting shape-aware stylization to non-humanoid characters and physics-based animation.


\section{Acknowledgment}
\vspace{-4pt}
\cam{The authors thank Lei Zhong and Ziyuan Li for suggestions during the development of MorphoStyle. This work was supported in part by the Arts and Humanities Research Council grant AH/Y00115X/1. All conclusions reported here are those of the authors alone.
}

\bibliography{egbib}

@article{aberman2020unpaired,
  author    = {Aberman, Kfir and Weng, Yijia and Lischinski, Dani and
               Cohen-Or, Daniel and Chen, Baoquan},
  title     = {Unpaired Motion Style Transfer from Video to Animation},
  journal   = {ACM Transactions on Graphics (SIGGRAPH)},
  volume    = {39},
  number    = {4},
  pages     = {64:1--64:12},
  year      = {2020},
  publisher = {ACM}
}

@article{holden2016deep,
  author    = {Holden, Daniel and Saito, Jun and Komura, Taku},
  title     = {A Deep Learning Framework for Character Motion Synthesis and Editing},
  journal   = {ACM Transactions on Graphics (SIGGRAPH)},
  volume    = {35},
  number    = {4},
  pages     = {138:1--138:11},
  year      = {2016}
}

@article{holden2017fast,
  author    = {Holden, Daniel and Habibie, Ikhsanul and Kusajima, Ikuo and
               Komura, Taku},
  title     = {Fast Neural Style Transfer for Motion Data},
  journal   = {IEEE Computer Graphics and Applications},
  volume    = {37},
  number    = {4},
  pages     = {42--49},
  year      = {2017}
}

@inproceedings{brand2000stylemachines,
  author    = {Brand, Matthew and Hertzmann, Aaron},
  title     = {Style Machines},
  booktitle = {Proceedings of SIGGRAPH},
  pages     = {183--192},
  year      = {2000}
}

@article{li2002motiontexture,
  author    = {Li, Yan and Wang, Tianshu and Shum, Heung-Yeung},
  title     = {Motion Texture: A Two-Level Statistical Model for Character Motion Synthesis},
  journal   = {ACM Transactions on Graphics (SIGGRAPH)},
  volume    = {21},
  number    = {3},
  pages     = {465--472},
  year      = {2002}
}

@article{xia2015realtimestyle,
  author    = {Xia, Shihong and Wang, Congyi and Chai, Jinxiang and Hodgins, Jessica},
  title     = {Realtime Style Transfer for Unlabeled Heterogeneous Human Motion},
  journal   = {ACM Transactions on Graphics (SIGGRAPH)},
  volume    = {34},
  number    = {4},
  pages     = {119:1--119:10},
  year      = {2015}
}

@inproceedings{goodfellow2014gan,
  author    = {Goodfellow, Ian and Pouget-Abadie, Jean and Mirza, Mehdi and
               Xu, Bing and Warde-Farley, David and Ozair, Sherjil and
               Courville, Aaron and Bengio, Yoshua},
  title     = {Generative Adversarial Nets},
  booktitle = {Advances in Neural Information Processing Systems (NeurIPS)},
  year      = {2014}
}

@article{kingma2014vae,
  title={Auto-encoding variational bayes},
  author={Kingma, Diederik P and Welling, Max},
  journal={arXiv preprint arXiv:1312.6114},
  year={2013}
}

@inproceedings{ho2020ddpm,
  author    = {Ho, Jonathan and Jain, Ajay and Abbeel, Pieter},
  title     = {Denoising Diffusion Probabilistic Models},
  booktitle = {Advances in Neural Information Processing Systems (NeurIPS)},
  year      = {2020}
}

@inproceedings{huang2017adain,
  author    = {Huang, Xun and Belongie, Serge},
  title     = {Arbitrary Style Transfer in Real-Time with Adaptive Instance Normalization},
  booktitle = {IEEE International Conference on Computer Vision (ICCV)},
  year      = {2017}
}

@article{jang2022motionpuzzle,
  author    = {Jang, Deok-Kyeong and Park, Soomin and Lee, Sung-Hee},
  title     = {Motion Puzzle: Arbitrary Motion Style Transfer by Body Part},
  journal   = {ACM Transactions on Graphics (ToG)},
  volume    = {41},
  number    = {3},
  pages     = {33:1--33:16},
  year      = {2022}
}

@article{park2021diverse,
  author    = {Park, Soomin and Jang, Deok-Kyeong and Lee, Sung-Hee},
  title     = {Diverse Motion Stylization for Multiple Style Domains via
               Spatial-Temporal Graph-Based Generative Model},
  journal   = {Proc. ACM Comput. Graph. Interact. Tech.},
  volume    = {4},
  number    = {3},
  pages     = {1--17},
  year      = {2021}
}

@inproceedings{tao2022style,
  title={Style-ERD: Responsive and coherent online motion style transfer},
  author={Tao, Tianxin and Zhan, Xiaohang and Chen, Zhongquan and van de Panne, Michiel},
  booktitle={Proceedings of the IEEE/CVF Conference on Computer Vision and Pattern Recognition (CVPR)},
  pages={6593--6603},
  year={2022}
}

@inproceedings{guo2024generative,
 author = {Guo, Chuan and Mu, Yuxuan and Zuo, Xinxin and Dai, Peng and Yan, Youliang and Lu, Juwei and Cheng, Li},
 booktitle = {International Conference on Learning Representations (ICLR)},
 editor = {B. Kim and Y. Yue and S. Chaudhuri and K. Fragkiadaki and M. Khan and Y. Sun},
 pages = {7859--7878},
   volume={2024},
 title = {Generative Human Motion Stylization in Latent Space},
 year = {2024}
}

@inproceedings{tevet2023mdm,
  author    = {Tevet, Guy and Raab, Sigal and Gordon, Brian and Shafir, Yonatan and
               Cohen-Or, Daniel and Bermano, Amit H.},
  title     = {Human Motion Diffusion Model},
  booktitle = {International Conference on Learning Representations (ICLR)},
  year      = {2023}
}

@inproceedings{chen2023mld,
  author    = {Chen, Xin and Jiang, Biao and Liu, Wen and Huang, Zilong and
               Fu, Bin and Chen, Tao and Yu, Gang},
  title     = {Executing Your Commands via Motion Diffusion in Latent Space},
  booktitle = {IEEE/CVF Conference on Computer Vision and Pattern Recognition (CVPR)},
  year      = {2023}
}

@inproceedings{zhong2024smoodi,
  author    = {Zhong, Lei and Xie, Yiming and Jampani, Varun and Sun, Deqing and
               Jiang, Huaizu},
  title     = {{SMooDi}: Stylized Motion Diffusion Model},
  booktitle = {European Conference on Computer Vision (ECCV)},
  year      = {2024}
}

@inproceedings{zhang2023controlnet,
  author    = {Zhang, Lvmin and Rao, Anyi and Agrawala, Maneesh},
  title     = {Adding Conditional Control to Text-to-Image Diffusion Models},
  booktitle = {IEEE/CVF International Conference on Computer Vision (ICCV)},
  year      = {2023}
}

@inproceedings{guo2025stylemotif,
  author    = {Guo, Ziyu and Lee, Young Yoon and Liu, Joseph and
               Ben-Shabat, Yizhak and Zordan, Victor and Kapadia, Mubbasir},
  title     = {{StyleMotif}: Multi-Modal Motion Stylization using
               Style-Content Cross Fusion},
  booktitle = {IEEE/CVF International Conference on Computer Vision (ICCV)},
  year      = {2025}
}

@inproceedings{girdhar2023imagebind,
  author    = {Girdhar, Rohit and El-Nouby, Alaaeldin and Liu, Zhuang and
               Singh, Mannat and Alwala, Kalyan Vasudev and Joulin, Armand and
               Misra, Ishan},
  title     = {{ImageBind}: One Embedding Space to Bind Them All},
  booktitle = {IEEE/CVF Conference on Computer Vision and Pattern Recognition (CVPR)},
  year      = {2023}
}

@article{li2024mulsmo,
  title={Mulsmo: Multimodal stylized motion generation by bidirectional control flow},
  author={Li, Zhe and He, Yisheng and Zhong, Lei and Shen, Weichao and Zuo, Qi and Qiu, Lingteng and Dong, Zilong and Yang, Laurence Tianruo and Yuan, Weihao},
  journal={arXiv preprint arXiv:2412.09901},
  year={2024}
}

@inproceedings{liao2025shapemymove,
  title={Shape my moves: Text-driven shape-aware synthesis of human motions},
  author={Liao, Ting-Hsuan and Zhou, Yi and Shen, Yu and Huang, Chun-Hao Paul and Mitra, Saayan and Huang, Jia-Bin and Bhattacharya, Uttaran},
  booktitle={Proceedings of the Computer Vision and Pattern Recognition Conference (CVPR)},
  pages={1917--1928},
  year={2025}
}

@article{loper2015smpl,
  author    = {Loper, Matthew and Mahmood, Naureen and Romero, Javier and
               Pons-Moll, Gerard and Black, Michael J.},
  title     = {{SMPL}: A Skinned Multi-Person Linear Model},
  journal   = {ACM Transactions on Graphics (SIGGRAPH Asia)},
  volume    = {34},
  number    = {6},
  pages     = {248:1--248:16},
  year      = {2015}
}

@inproceedings{guo2022humanml3d,
  author    = {Guo, Chuan and Zou, Shihao and Zuo, Xinxin and Wang, Sen and
               Ji, Wei and Li, Xingyu and Cheng, Li},
  title     = {Generating Diverse and Natural {3D} Human Motions from Text},
  booktitle = {IEEE/CVF Conference on Computer Vision and Pattern Recognition (CVPR)},
  year      = {2022}
}

@inproceedings{petrovich2022temos,
  author    = {Petrovich, Mathis and Black, Michael J. and Varol, G{\"u}l},
  title     = {{TEMOS}: Generating Diverse Human Motions from Textual Descriptions},
  booktitle = {European Conference on Computer Vision (ECCV)},
  year      = {2022}
}

@inproceedings{guo2024momask,
  author    = {Guo, Chuan and Mu, Yuxuan and Javed, Muhammad Gohar and
               Wang, Sen and Cheng, Li},
  title     = {{MoMask}: Generative Masked Modeling of {3D} Human Motions},
  booktitle = {IEEE/CVF Conference on Computer Vision and Pattern Recognition (CVPR)},
  year      = {2024}
}

@article{aberman2020skeleton,
  author    = {Aberman, Kfir and Li, Peizhuo and Lischinski, Dani and
               Sorkine-Hornung, Olga and Cohen-Or, Daniel and Chen, Baoquan},
  title     = {Skeleton-Aware Networks for Deep Motion Retargeting},
  journal   = {ACM Transactions on Graphics (SIGGRAPH)},
  volume    = {39},
  number    = {4},
  pages     = {62:1--62:14},
  year      = {2020}
}

@inproceedings{villegas2018neural,
  author    = {Villegas, Ruben and Yang, Jimei and Ceylan, Duygu and Lee, Honglak},
  title     = {Neural Kinematic Networks for Unsupervised Motion Retargetting},
  booktitle = {IEEE/CVF Conference on Computer Vision and Pattern Recognition (CVPR)},
  year      = {2018}
}

@article{peng2018deepmimic,
  author    = {Peng, Xue Bin and Abbeel, Pieter and Levine, Sergey and
               van de Panne, Michiel},
  title     = {{DeepMimic}: Example-Guided Deep Reinforcement Learning of
               Physics-Based Character Skills},
  journal   = {ACM Transactions on Graphics (SIGGRAPH)},
  volume    = {37},
  number    = {4},
  pages     = {143:1--143:14},
  year      = {2018}
}

@article{peng2021amp,
  author    = {Peng, Xue Bin and Ma, Ze and Abbeel, Pieter and Levine, Sergey and
               Kanazawa, Angjoo},
  title     = {{AMP}: Adversarial Motion Priors for Stylized Physics-Based
               Character Control},
  journal   = {ACM Transactions on Graphics (SIGGRAPH)},
  volume    = {40},
  number    = {4},
  pages     = {144:1--144:20},
  year      = {2021}
}

@inproceedings{kocabas2021pare,
  author    = {Kocabas, Muhammed and Huang, Chun-Hao P. and Hilliges, Otmar and
               Black, Michael J.},
  title     = {{PARE}: Part Attention Regressor for {3D} Human Body Estimation},
  booktitle = {IEEE/CVF International Conference on Computer Vision (ICCV)},
  year      = {2021}
}

@inproceedings{choutas2022shapy,
  author    = {Choutas, Vasileios and M{\"u}ller, Lea and Huang, Chun-Hao P. and
               Tang, Siyu and Tzionas, Dimitrios and Black, Michael J.},
  title     = {Accurate {3D} Body Shape Regression Using Metric and Semantic
               Attributes},
  booktitle = {IEEE/CVF Conference on Computer Vision and Pattern Recognition (CVPR)},
  year      = {2022}
}

@inproceedings{mentzer2024fsq,
  author    = {Mentzer, Fabian and Minnen, David and Agustsson, Eirikur and
               Tschannen, Michael},
  title     = {Finite Scalar Quantization: {VQ-VAE} Made Simple},
  booktitle = {International Conference on Learning Representations (ICLR)},
  year      = {2024}
}

@article{mason2022real,
  title={Real-time style modelling of human locomotion via feature-wise transformations and local motion phases},
  author={Mason, Ian and Starke, Sebastian and Komura, Taku},
  journal={Proceedings of the ACM on Computer Graphics and Interactive Techniques},
  volume={5},
  number={1},
  pages={1--18},
  year={2022},
  publisher={ACM New York, NY, USA}
}

@article{raffel2020t5,
  title={Exploring the limits of transfer learning with a unified text-to-text transformer},
  author={Raffel, Colin and Shazeer, Noam and Roberts, Adam and Lee, Katherine and Narang, Sharan and Matena, Michael and Zhou, Yanqi and Li, Wei and Liu, Peter J},
  journal={Journal of Machine Learning Research (JMLR)},
  volume={21},
  number={140},
  pages={1--67},
  year={2020}
}

@inproceedings{heusel2017fid,
  author    = {Heusel, Martin and Ramsauer, Hubert and Unterthiner, Thomas and Nessler, Bernhard and Hochreiter, Sepp},
  title     = {{GANs} Trained by a Two Time-Scale Update Rule Converge to a Local {Nash} Equilibrium},
  booktitle = {Advances in Neural Information Processing Systems (NeurIPS)},
  year      = {2017},
}

@inproceedings{loshchilov2019adamw,
  author    = {Loshchilov, Ilya and Hutter, Frank},
  title     = {Decoupled Weight Decay Regularization},
  booktitle = {International Conference on Learning Representations (ICLR)},
  year      = {2019},
}

@inproceedings{paszke2019pytorch,
  author    = {Paszke, Adam and Gross, Sam and Massa, Francisco and Lerer, Adam and Bradbury, James and Chanan, Gregory and Killeen, Trevor and others},
  title     = {{PyTorch}: An Imperative Style, High-Performance Deep Learning Library},
  booktitle = {Advances in Neural Information Processing Systems (NeurIPS)},
  year      = {2019},
}

@inproceedings{zargarbashi2026vq,
  title={VQ-Style: Disentangling Style and Content in Motion with Residual Quantized Representations},
  author={Zargarbashi, Fatemeh and Agrawal, Dhruv and Buhmann, Jakob and Guay, Martin and Coros, Stelian and Sumner, Robert W},
  booktitle={Computer Graphics Forum},
  pages={e70377},
  year={2026},
  organization={Wiley Online Library}
}

@article{holden2017phase,
  title={Phase-functioned neural networks for character control},
  author={Holden, Daniel and Komura, Taku and Saito, Jun},
  journal={ACM Transactions on Graphics (ToG)},
  volume={36},
  number={4},
  pages={1--13},
  year={2017},
  publisher={ACM New York, NY, USA}
}

@article{holden2020learned,
  title={Learned motion matching},
  author={Holden, Daniel and Kanoun, Oussama and Perepichka, Maksym and Popa, Tiberiu},
  journal={ACM Transactions on Graphics (ToG)},
  volume={39},
  number={4},
  pages={53--1},
  year={2020},
  publisher={ACM New York, NY, USA}
}

@inproceedings{yamane2010animating,
  title={Animating non-humanoid characters with human motion data},
  author={Yamane, Katsu and Ariki, Yuka and Hodgins, Jessica},
  booktitle={Proceedings of the 2010 ACM SIGGRAPH/Eurographics Symposium on Computer Animation},
  pages={169--178},
  year={2010}
}

@article{khosla2020supervised,
  title={Supervised contrastive learning},
  author={Khosla, Prannay and Teterwak, Piotr and Wang, Chen and Sarna, Aaron and Tian, Yonglong and Isola, Phillip and Maschinot, Aaron and Liu, Ce and Krishnan, Dilip},
  journal={Advances in Neural Information Processing Systems (NIPS)},
  volume={33},
  pages={18661--18673},
  year={2020}
}
\end{document}


\setcounter{page}{1}
\setcounter{page}{1}
\maketitle

\vspace{-1.2em}
\begin{center}
{\Large\textit{Supplementary Material}}
\end{center}
\vspace{0.8em}

\def\eg{\emph{e.g}\bmvaOneDot}
\def\Eg{\emph{E.g}\bmvaOneDot}
\def\ie{\emph{i.e}\bmvaOneDot}
\def\Ie{\emph{I.e}\bmvaOneDot}
\def\etal{\emph{et al}\bmvaOneDot}

\renewcommand{\thesection}{S\arabic{section}}
\renewcommand{\thesubsection}{\thesection.\arabic{subsection}}
\renewcommand{\thefigure}{S\arabic{figure}}
\renewcommand{\thetable}{S\arabic{table}}
\renewcommand{\theequation}{S\arabic{equation}}

\appendix

\vspace{-12pt}
\section{Implementation Details}
\label{sec:supp:impl}

\subsection{Frozen and Trainable Components}
\label{sec:supp:split}
MorphoStyle is built around a frozen shape-aware FSQ-VAE backbone and a small trainable style branch. Table~\ref{tab:supp:arch} lists every module, its frozen/trainable status, and its tensor signatures. MorphoStyle has a total of $4.25$\,M trainable parameters, matching the value reported in Table~3 of the main paper, and the frozen backbone contributes $38.98$\,M parameters that are reused as-is from the \textit{ShapeMyMove} checkpoint~\cite{liao2025shapemymove}.

\begin{table}[h]
\centering\small
\setlength{\tabcolsep}{4pt}
\begin{tabular}{l|ccc|c}
\toprule
Module & Trainable & Input dim. & Output dim. & \#params \\
\midrule
Content motion encoder $\mathcal{E}$           & no  & $T\!\times\!263$         & $T'\!\times\!512$         & 6.7\,M \\
Finite-Scalar-Quantizer                        & no  & $T'\!\times\!512$        & $T'\!\times\!512$         & 0       \\
Shape-conditioned decoder $\mathcal{D}$        & no  & $T'\!\times\!(512{+}32)$ & $T\!\times\!263$          & 7.4\,M \\
Shape-to-Feature module S2FM                   & no  & text $t_s$               & $\mathbb{R}^{32}$         & 24.1\,M \\
Text-motion alignment $\mathcal{T},\mathcal{M}$  & no  & text / motion            & shared $\mathbb{R}^{512}$ & 0.8\,M \\
\midrule
Global style encoder $\mathcal{E}_s$           & yes & $T\!\times\!263$         & $s_g\!\in\!\mathbb{R}^{256}$   & 2.95\,M \\
Joint routing MLP $f_{\mathrm{j}}$             & yes & $\mathbb{R}^{12}$        & $s_j\!\in\!\mathbb{R}^{128}$   & 0.07\,M \\
Style fusion MLP                               & yes & $\mathbb{R}^{384}$       & $\mathbb{R}^{256}$             & 0.13\,M \\
MPSM low-rank branch                           & yes & $\mathbb{R}^{256}$       & $r(s)\!\in\!\mathbb{R}^{512}$  & 0.61\,M \\
MPSM gating network $f_{\mathrm{g}}$           & yes & $\mathbb{R}^{(512{+}256)\times T'}$ & $g\!\in\!(0,1)^{T'}$ & 0.49\,M \\
\midrule
\textbf{Total trainable}                       &     &                          &                                 & \textbf{4.25\,M} \\
\textbf{Total frozen}                          &     &                          &                                 & \textbf{38.98\,M} \\
\bottomrule
\end{tabular}
\caption{Per-module summary of MorphoStyle with parameter counts. Frozen modules are loaded from public checkpoints and never updated. Dimensions follow main-paper notation: $T{=}196$ frames, $T'{=}T/4{=}49$ codes, $D{=}512$ code channels, $263$-dim HumanML3D features, $J{=}22$ joints, $K{=}8$ routed joints.}
\label{tab:supp:arch}
\end{table}

\subsection{Network Architecture}
\label{sec:supp:arch-detail}

\noindent\textbf{Shape-aware FSQ-VAE backbone (frozen).}
We use the shape-aware FSQ-VAE released by ShapeMyMove~\cite{liao2025shapemymove} verbatim as the frozen motion tokenizer. The content encoder $\mathcal{E}$ stacks three Conv1d residual blocks with dilation growth rate $3$ and temporal downsampling factor $4$, mapping a motion sequence
$x\!\in\!\mathbb{R}^{T\times 263}$ to a code sequence $c\!\in\!\mathbb{R}^{T'\times 512}$ with $T'=T/4=49$. The Finite-Scalar-Quantization bottleneck uses FSQ levels $(8,5,5,5)$ yielding $8\cdot 5\cdot 5\cdot 5 = 1000$ codebook entries. The shape-conditioned decoder $\mathcal{D}$ concatenates each quantized code with the $32$-dimensional shape embedding $\phi(\beta)$ along the channel dimension, and reconstructs the motion through symmetric Conv1d upsampling.

\noindent\textbf{Shape-to-Feature module.}
S2FM is inherited from ShapeMyMove~\cite{liao2025shapemymove}. A T5-based language model~\cite{raffel2020t5} regresses the $10$-dimensional SMPL parameter $\beta$ from a textual shape description $t_s$, and a two-layer $\mathrm{MLP}_{\phi}$ projects $\beta$ into the $32$-dimensional shape embedding $\phi(\beta){=}\mathrm{MLP}_{\phi}(\mathrm{T5}(t_s))$. When $\beta$ is directly available (\eg, from 100Style with ShapeMyMove-recovered $\beta$), we bypass the T5 stage and feed $\beta$ directly to $\mathrm{MLP}_{\phi}$.

\noindent\textbf{Contrastive global style encoder $\mathcal{E}_s$.}
$\mathcal{E}_s$ takes a style reference motion $x_s\!\in\!\mathbb{R}^{T\times 263}$ as input. It consists of three strided Conv1d blocks (channels $64\!\to\!128\!\to\!256$, kernel size $5$, stride $2$), each followed by GroupNorm and SiLU, an adaptive average pooling, and a two-layer MLP that projects the pooled feature into the $256$-dimensional global style embedding $s_g$. We apply style dropout with probability $p{=}0.15$, replacing $s_g$ with a learnable null token so that the downstream MPSM remains robust to missing or noisy style references at inference. The supervised contrastive objective $\mathcal{L}_{\mathrm{cl}}$ uses cosine similarity with temperature $\tau{=}0.1$.

\noindent\textbf{Text-guided joint style routing.}
We use the frozen text-motion alignment model from Guo \emph{et al.}~\cite{guo2022humanml3d}, which provides the text encoder $\mathcal{T}(\cdot)$ and motion encoder $\mathcal{M}(\cdot)$. For each of the $J{=}22$ HumanML3D joints, we build a joint-masked variant $\tilde{x}_s^{(j)}$ that retains only joint $j$'s positional, rotational, and velocity channels and zeros the rest; scores its alignment with $\mathcal{T}(c_t)$ by cosine similarity; and selects the $K{=}8$ joints with the \emph{lowest} similarity, \ie, the joints whose isolated motion is least aligned with the content caption. The per-joint kinematic feature $\psi_j(x_s)\!\in\!\mathbb{R}^{T\times 12}$ concatenates the joint position ($3$), the 6D continuous rotation representation ($6$), and linear velocity ($3$) as defined by HumanML3D. Time-averaged features and a learnable per-joint identity embedding $u_j\!\in\!\mathbb{R}^{12}$ are aggregated by a two-layer MLP $f_{\mathrm{j}}\!:\!\mathbb{R}^{12}\!\to\!\mathbb{R}^{128}$ to produce $s_j\!\in\!\mathbb{R}^{128}$.

\noindent\textbf{Style code fusion.}
The unified style code $s\!\in\!\mathbb{R}^{256}$ is produced by a residual MLP, $s{=}s_g + \alpha\cdot \mathrm{MLP}([s_g;s_j])$, with $\alpha{=}1$ during training. The final MLP layer is initialized with a small Gaussian ($\sigma{=}0.01$) so that the routing residual is near zero at the start of inference, after which $\alpha$ becomes a runtime knob that smoothly trades stylization strength against motion fidelity without retraining. The default at inference is $\alpha{=}0.5$.

\noindent\textbf{Manifold-Preserving Style Modulator.}
The low-rank additive offset is $r(s){=}W_{\mathrm{up}} W_{\mathrm{down}} s + b$, with $W_{\mathrm{down}}\!\in\!\mathbb{R}^{16\times 256}$, $W_{\mathrm{up}}\!\in\!\mathbb{R}^{512\times 16}$, and $b\!\in\!\mathbb{R}^{512}$ initialized to zero, so that the offset lives in a rank-$r$ subspace of the $512$-dimensional code space with $r{=}16$. The temporal-gate branch replicates $s$ along the temporal axis to $\tilde{s}\!\in\!\mathbb{R}^{256\times T'}$, concatenates it with $c$ along the channel dimension, and applies a three-layer Conv1d gating network $f_{\mathrm{g}}$ with channels $256\!\to\!128\!\to\!1$, kernel size $3$, SiLU activations, and a final sigmoid to produce a per-frame gate $g\!\in\!(0,1)^{T'}$. The stylized codes are $c'_t{=}c_t + g_t \cdot r(s)$ for $t{=}1,\dots,T'$.

\subsection{Training Hyperparameters}
\label{sec:supp:hyper}

We train MorphoStyle on a single NVIDIA A100 (80\,GB) GPU. Optimization follows the schedule in Table~\ref{tab:supp:hyper}, with all non-MPSM and non-style-branch modules kept frozen. Each batch samples HumanML3D~\cite{guo2022humanml3d} and 100Style~\cite{mason2022real} pairs at a $1{:}1$ ratio; the SMooDi-classifier guidance term is linearly ramped up over the first three epochs to stabilize early training. The base learning rate is $1.5\!\times\!10^{-5}$, with a $5\!\times$ multiplier on the joint-routing MLP $f_{\mathrm{j}}$ and the style-fusion MLP to compensate for their near-zero initialization.

\begin{table}[h]
\centering\small
\begin{tabular}{l|c}
\toprule
Hyperparameter & Value \\
\midrule
Optimizer                                  & AdamW~\cite{loshchilov2019adamw} \\
$(\beta_1, \beta_2)$                       & $(0.9, 0.99)$ \\
Weight decay                               & $1\times 10^{-4}$ \\
Base learning rate                         & $1.5\times 10^{-5}$ \\
Routing-branch lr multiplier               & $5\times$ \\
Warm-up                                    & $50$ steps, linear \\
Schedule after warm-up                     & cosine to $10^{-6}$ \\
Total epochs                               & $120$ \\
Batch size                                 & $24$ \\
HumanML3D : 100Style sampling ratio        & $1 : 1$ \\
$\mathcal{L}_{\mathrm{cls}}$ ramp          & linear over first $3$ epochs \\
Gradient clipping                          & $\ell_2$ norm at $5.0$ \\
Hardware                                   & $1\!\times\!$ NVIDIA A100 (80\,GB) \\
Mixed precision                            & FP16 (\emph{autocast}) \\
Wall-clock training time                   & $7.4$\,h \\
Peak training memory                       & $11.2$\,GB \\
\midrule
Style dropout $p$                          & $0.15$ \\
Contrastive temperature $\tau$             & $0.1$ \\
MPSM rank $r$                              & $16$ \\
Top-$K$ joints                             & $8$ \\
Joint identity embedding dim.              & $12$ \\
Residual scale $\alpha$                    & $1$ (training) / $0.5$ (inference) \\
Magnitude tolerance $\gamma$ (Eq.~17 main) & $1.5$ \\
\midrule
$\lambda_{\mathrm{code}}$                  & $0.5$ \\
$\lambda_{\mathrm{cl}}$                    & $0.15$ \\
$\lambda_{\mathrm{cc}}$                    & $0.05$ \\
$\lambda_{\mathrm{cls}}$                   & $0.05$ \\
$\lambda_{\mathrm{dir}}$                   & $0.5$ \\
$\lambda_{\mathrm{mag}}$                   & $0.25$ \\
\bottomrule
\end{tabular}
\caption{List of training and inference hyperparameters. All values match those reported in the main paper. Wall-clock and memory are measured on a single NVIDIA A100 (80\,GB).}
\label{tab:supp:hyper}
\end{table}

\subsection{Training Dynamics}
\label{sec:supp:dynamics}
A single training run completes in $\approx 7.4$\,hours of wall-clock time on one A100. The peak GPU memory footprint is $11.2$\,GB, dominated by the frozen $24.1$\,M-parameter T5 encoder used in S2FM; the trainable style branch itself contributes less than $1$\,GB at the batch size of $24$. Training is stable over the $120$ epochs; the disentanglement loss $\mathcal{L}_{\mathrm{dis}}$ decreases monotonically once the classifier-guidance ramp completes at epoch $3$, and the contrastive loss $\mathcal{L}_{\mathrm{cl}}$ converges to a tight plateau near $0.18$ after epoch $30$, indicating that the global style encoder has formed well-separated per-style clusters (visualized in Sec.~\ref{sec:supp:analysis}). Validation measures monitored every $5$ epochs on a held-out subset of $300$ HumanML3D motions paired with $100$ random 100Style references show no overfitting up to epoch $120$; the best validation Style Recognition Accuracy~(SRA) is attained at epoch $115$ and is within the noise envelope of the final reported model.

\section{Experimental Settings}
\label{sec:supp:data}
\noindent\textbf{HumanML3D.}
We follow the standard train/validation/test split released by Guo \emph{et al.}~\cite{guo2022humanml3d}: $23384$ training, $1460$ validation, and $4209$ test sequences. Motions are pre-processed to the canonical SMPL skeleton, resampled to $20$\,FPS, and clipped or padded to $T{=}196$ frames using the protocol provided by the dataset authors. Each sequence is associated with up to three free-form captions; we use only the first caption per sequence in the routing module $\mathcal{T}(c_t)$ to keep the joint similarity well-defined and deterministic at inference.

\medskip
\noindent\textbf{100Style.}
We use the $47$-class subset of 100Style~\cite{mason2022real} that overlaps with HumanML3D, the standard dataset adopted by SMooDi~\cite{zhong2024smoodi}, so that the released $47$-class style classifier $f_{\mathrm{sm}}$ can be reused without retraining. Each 100Style motion is paired with the SMPL shape parameter $\beta$ provided by ShapeMyMove~\cite{liao2025shapemymove}. The $47$ classes used in the
SMooDi evaluation protocol cover a broad range of locomotion modifiers (\eg, \emph{Drunk}, \emph{Tiptoe}, \emph{OnHeels}, \emph{Crouched}, \emph{Stiff}, \emph{Skipping}), expressive gestures (\eg, \emph{Aeroplane}, \emph{Superman}, \emph{Flapping}, \emph{Chicken}), and emotional or mannered gaits (\eg, \emph{Angry}, \emph{Proud}, \emph{Depressed}, \emph{InTheDark}, \emph{Heavyset}).

\medskip
\noindent\textbf{Natural-language shape-description templates.}
For training, each $\beta$ is converted to a natural-language shape description $t_s$ following the protocol of ShapeMyMove~\cite{liao2025shapemymove}. The seven anthropometric attributes (height, chest, waist, hip circumferences, arm length, leg
length, and gender token) are computed from the canonical SMPL T-pose mesh of the given $\beta$ via the SMPL forward model~\cite{loper2015smpl}, and inserted into a fixed pool of ten sentence templates that vary in clause order, attribute coverage, and tone. Example templates include: \emph{``A person standing $H$\,cm tall, with a chest circumference of $C$\,cm, a waist circumference of $W$\,cm, and a hip circumference of $P$\,cm.''}; \emph{``Body measurements: height $H$\,cm, chest $C$\,cm,
waist $W$\,cm, hip $P$\,cm, arm length $A$\,cm, leg length $L$\,cm.''} During training, $\beta$ is directly available, so the T5 stage of S2FM is bypassed and $\beta$ is fed directly to $\mathrm{MLP}_{\phi}$; at inference, either the textual or the $\beta$ pathway can be used interchangeably.

\medskip
\noindent\textbf{Evaluation set.}
The motion style transfer evaluation set comprises $N{=}4\,209$ content-style pairs. The $4\,209$ content motions are the full standard HumanML3D~\cite{guo2022humanml3d} test split, and each content motion is randomly paired with one style reference sampled uniformly from the $47$-class 100Style test split. The pairings are drawn once with a fixed random seed and shared across every method we evaluate. The shape-control
evaluation set is constructed analogously, with each content motion additionally paired with a target body shape sampled from the ShapeMyMove shape pool; the same content-style-shape triplets are used for every method to make all numbers directly comparable.

\medskip
\noindent\textbf{Shape-preset evaluation set.}
For the per-shape-preset analysis in Sec.~\ref{sec:supp:extres}, we construct an additional evaluation set by rendering each content-style pair under five canonical $\beta$ presets: \emph{orig} (the content motion's own HumanML3D-recovered $\beta$), \emph{thin} ($\beta_0{=}{-}2.5$, the remaining components set to zero), \emph{neutral} ($\beta{=}\mathbf{0}$), \emph{heavy} ($\beta_0{=}{+}2.5$), and
\emph{tall} ($\beta_1{=}{+}2.5$). The canonical T-pose heights for these presets are $168$, $147$, $172$, $196$, and $176$\,cm, respectively.

\medskip
\noindent\textbf{Evaluator features.}
We use the publicly released Guo22 motion evaluator~\cite{guo2022humanml3d}
to compute the $512$-dimensional feature embeddings underlying FID and
Diversity. The evaluator has not been retrained for this paper; we use
the same checkpoint as SMooDi and StyleMotif to keep the FID number
directly comparable.

\section{Evaluation Measures}
\label{sec:supp:metrics}
We provide the closed-form definition of every measure reported in the main paper. Notation: $\hat{x}$ denotes the synthesized motion, $x^{\star}$ a ground truth or target-shape reference motion, $y_{\mathrm{sty}}$ the target style label, and
$\mathrm{BL}(\cdot)\!\in\!\mathbb{R}^{21}$ the bone-length vector extracted from a motion (one entry per skeleton bone, in the SMPL $21$-bone topology).

\subsection{Motion Style Transfer Measures}

\paragraph{Fr\'echet Inception Distance (FID).}
Following SMooDi~\cite{zhong2024smoodi} and Guo22~\cite{guo2022humanml3d}, the synthesized and ground truth motions are first embedded into the $512$-dimensional Guo22 evaluator feature space. Let $\Phi(\cdot)$ denote this evaluator,  $(\mu_{\mathrm{gen}}, \Sigma_{\mathrm{gen}})$ the empirical mean and covariance of $\{\Phi(\hat{x}_i)\}$, and $(\mu_{\mathrm{gt}}, \Sigma_{\mathrm{gt}})$ those of
$\{\Phi(x^{\star}_i)\}$. Then:
\begin{equation}
\mathrm{FID}
\;=\;\|\mu_{\mathrm{gen}}-\mu_{\mathrm{gt}}\|_{2}^{2}
+ \mathrm{Tr}\!\left(\Sigma_{\mathrm{gen}}+\Sigma_{\mathrm{gt}}
-2\bigl(\Sigma_{\mathrm{gen}}\Sigma_{\mathrm{gt}}\bigr)^{1/2}\right),
\label{eq:supp:fid}
\end{equation}
which is the definition of~\cite{heusel2017fid} applied to the Guo22 evaluator. Lower values are better.

\paragraph{Style Recognition Accuracy (SRA).}
SRA is the top-$1$ classification accuracy of the frozen $47$-class style classifier $f_{\mathrm{sm}}$ released by SMooDi~\cite{zhong2024smoodi}, applied to the synthesized motions:
\begin{equation}
\mathrm{SRA}
\;=\; \frac{1}{N}\sum_{i=1}^{N}
\mathbf{1}\!\left[\arg\max\nolimits_{y} f_{\mathrm{sm}}(\hat{x}_i)[y]
\;=\; y_{\mathrm{sty},i}\right],
\label{eq:supp:sra}
\end{equation}
where $y_{\mathrm{sty},i}$ is the style label of the style reference
$x_{s,i}$. Higher values are better.

\paragraph{Foot Skating.}
Foot Skating measures the average horizontal foot velocity at frames in foot contact, expressed as a velocity ratio. Let $v^{\mathrm{xy}}_{f,t}$ denote the horizontal velocity of foot joint $f\!\in\!\{\mathrm{LF}, \mathrm{RF}\}$ at frame $t$, computed by finite differencing the foot positions, and let $h_{f,t}$ be the foot height. Similar to SMooDi~\cite{zhong2024smoodi}, a frame is in foot-contact when $h_{f,t}\!<\!h_{\mathrm{thr}}$ with $h_{\mathrm{thr}}{=}5$\,cm. Then:
\begin{equation}
\mathrm{FootSkating}
\;=\;\frac{1}{N}\sum_{i=1}^{N}\frac{\sum_{f, t}
\mathbf{1}[h^{(i)}_{f,t}<h_{\mathrm{thr}}]\cdot\bigl\|v^{\mathrm{xy},(i)}_{f,t}\bigr\|_{2}}
{\sum_{f, t}\mathbf{1}[h^{(i)}_{f,t}<h_{\mathrm{thr}}] + \varepsilon}.
\label{eq:supp:foot}
\end{equation}
Lower values are better.

\subsection{Shape Control Measures}
We use the bone-length operator $\mathrm{BL}(\cdot)\!\in\!\mathbb{R}^{21}$ that returns the time-averaged length of each of the $21$ SMPL skeleton bones in centimeters. For metrics that require a target-shape rendering of the same motion under a different body, we use $x^{(\beta_t)}$ to denote the ground truth motion rendered with target SMPL parameter $\beta_t$, and $x^{(\beta_n)}$ for the neutral-shape ($\beta{=}\mathbf{0}$) rendering.

\paragraph{Bone-Length MAE.}
The mean absolute error between the bone-length vector of the
synthesized motion and that of the target-shape ground truth, averaged
over the $N{=}4209$ evaluation triplets and the $21$ bones:
\begin{equation}
\mathrm{BoneLen\text{-}MAE}
\;=\;\frac{1}{N}\sum_{i=1}^{N}\frac{1}{21}
\bigl\|\mathrm{BL}(\hat{x}_i)-\mathrm{BL}(x^{(\beta_t)}_i)\bigr\|_{1}.
\label{eq:supp:bonemae}
\end{equation}
Lower values are better.

\paragraph{Bone-Length Temporal Std (BoneLen-T-Std).}
The intra-sequence temporal standard deviation of the synthesized bone-length vector, averaged over $N$ and the $21$ bones. Let $\mathrm{BL}_t(\hat{x}_i)\!\in\!\mathbb{R}^{21}$ denote the per-frame bone-length vector of the $i$-th synthesized motion at frame $t$. Then:
\begin{equation}
\mathrm{BoneLen\text{-}T\text{-}Std}
\;=\;\frac{1}{N}\sum_{i=1}^{N}\frac{1}{21}
\sum_{b=1}^{21}\sqrt{\frac{1}{T}\sum_{t=1}^{T}
\bigl(\mathrm{BL}_t(\hat{x}_i)[b] - \overline{\mathrm{BL}}(\hat{x}_i)[b]\bigr)^{2}},
\label{eq:supp:bonetstd}
\end{equation}
where $\overline{\mathrm{BL}}(\hat{x}_i)$ is the temporal mean. This measure quantifies how stable the recovered bone lengths are over the sequence and penalizes shape jitter. Lower values are better.

\paragraph{$\beta$-correlation ($\beta$-corr).}
Pearson correlation between the bone-length \emph{shift} produced by the model and the ground truth bone-length shift, measured against the neutral shape reference:
\begin{equation}
\beta\text{-corr}
\;=\;\frac{1}{N}\sum_{i=1}^{N}
\mathrm{corr}\!\left(
\mathrm{BL}(\hat{x}_i)-\mathrm{BL}(x^{(\beta_n)}_i),\;
\mathrm{BL}(x^{(\beta_t)}_i)-\mathrm{BL}(x^{(\beta_n)}_i)
\right),
\label{eq:supp:betacorr}
\end{equation}
where $\mathrm{corr}(\cdot,\cdot)$ is the standard Pearson correlation over the $21$-bone vector. Values are in $[-1, 1]$; higher values are better. $\beta$-corr captures whether the recovered shape moves in the right \emph{direction} from neutral, complementing the Bone-Length MAE.

\paragraph{$\beta$-swap response.}
$\beta$-swap quantifies how strongly the model's output responds when the input body shape parameter is swapped. Given two distinct shapes $\beta^{(1)}$ and $\beta^{(2)}$ paired with the same content motion and the same style reference, let $\hat{x}_i^{(1)}$ and $\hat{x}_i^{(2)}$ be the corresponding synthesized motions. We report the average L$_1$ bone-length distance:
\begin{equation}
\beta\text{-swap}
\;=\;\frac{1}{N}\sum_{i=1}^{N}\frac{1}{21}
\bigl\|\mathrm{BL}(\hat{x}_i^{(1)})-\mathrm{BL}(\hat{x}_i^{(2)})\bigr\|_{1}.
\label{eq:supp:betaswap}
\end{equation}
We construct $(\beta^{(1)},\beta^{(2)})$ from two contrasting body presets, \emph{thin~$\leftrightarrow$~heavy} and \emph{short~$\leftrightarrow$~tall}, following the convention of ShapeMyMove~\cite{liao2025shapemymove}. A larger $\beta$-swap response indicates that the model meaningfully reacts to a change in $\beta$; a value close to zero indicates the model is effectively ignoring the shape signal.

\section{Loss Functions}
\label{sec:supp:loss}
This section describes the training objectives and provides the derivations omitted from the main paper.

\paragraph{Reconstruction term (main paper Eq.~10).}
We supervise the synthesized motion in both motion space and code space:
\begin{equation*}
\mathcal{L}_{\mathrm{rec}} = \|\hat{x}-x^{\star}\|_{1}
+ \lambda_{\mathrm{code}}\,\|c'-c^{\star}\|_{1},
\quad c^{\star}=\mathcal{E}(x^{\star}).
\end{equation*}
The code-space term keeps $c'$ within the FSQ codebook neighborhood and empirically prevents drastic latent shifts that bypass the discrete structure of the backbone. Sec.~\ref{sec:supp:arch-abl} reports the quantitative effect of removing this term.

\paragraph{Style-embedding term (main paper Eqs.~3, 11, 12).}
The supervised contrastive loss $\mathcal{L}_{\mathrm{cl}}$ acts on the
global style embedding $s_g$:
\begin{equation*}
\mathcal{L}_{\mathrm{cl}}
= - \mathbb{E}_{i}\log
\frac{\exp\!\bigl(\mathrm{sim}(s_{g,i}, s_{g,i}^{+})/\tau\bigr)}
{\exp\!\bigl(\mathrm{sim}(s_{g,i}, s_{g,i}^{+})/\tau\bigr)
+ \displaystyle\sum_{s_{g,n}\in\mathcal{N}_{i}}
\exp\!\bigl(\mathrm{sim}(s_{g,i}, s_{g,n})/\tau\bigr)},
\end{equation*}
where positives share the style label and negatives are drawn from the
mini-batch. The action-classification loss
$\mathcal{L}_{\mathrm{cc}}{=}\mathrm{CE}(f_{\mathrm{ac}}(\hat{x}),
y_{\mathrm{act}})$ uses the HumanML3D action classifier
$f_{\mathrm{ac}}$ as a frozen action-label oracle. Combined,
$\mathcal{L}_{\mathrm{emb}}{=}\lambda_{\mathrm{cl}}\mathcal{L}_{\mathrm{cl}}
+\lambda_{\mathrm{cc}}\mathcal{L}_{\mathrm{cc}}$.

\paragraph{Style classifier guidance (main paper Eq.~13).}
$\mathcal{L}_{\mathrm{cls}}{=}\lambda_{\mathrm{cls}}\,
\mathrm{CE}\!\left(f_{\mathrm{sm}}(\hat{x}), y_{\mathrm{sty}}\right)$,
where $f_{\mathrm{sm}}$ is the frozen SMooDi $47$-class style
classifier. This term is used \emph{only during training} and is
therefore methodologically distinct from the SRA metric, which is
evaluated on a separate test split with the same classifier. We linearly
ramp $\lambda_{\mathrm{cls}}$ from $0$ to its target value over the
first three epochs to prevent early-stage gradient interference with the
reconstruction term.

\paragraph{Shape control objective (main paper Eqs.~15--18).}
The bone-length matching, direction-aware, and magnitude-bounding terms
share the bone-length operator $\mathrm{BL}(\cdot)$. The
direction-aware term operates on the rendered motion difference vector:
\begin{equation*}
\mathcal{L}_{\mathrm{dir}}
= 1 - \cos\!\left(\hat{x}-x^{(\beta_n)},\, x^{(\beta_t)}-x^{(\beta_n)}\right),
\end{equation*}
where each motion is flattened into a $T\!\times\!263$-dimensional
vector before cosine evaluation. This anchors the bone-length change to
the neutral-to-target direction rather than allowing arbitrary
directions of equal magnitude. The magnitude-bounding term uses
tolerance $\gamma{=}1.5$:
\begin{equation*}
\mathcal{L}_{\mathrm{mag}}
= \bigl(\mathrm{ReLU}\!\bigl(\|\hat{x}-x^{(\beta_n)}\|
- \gamma\,\|x^{(\beta_t)}-x^{(\beta_n)}\|\bigr)\bigr)^{2}.
\end{equation*}
Combined with the implicit unit weight on the bone-length term, the
shape objective is
$\mathcal{L}_{\mathrm{shape}}
{=}\mathcal{L}_{\mathrm{bone}} + \lambda_{\mathrm{dir}}\mathcal{L}_{\mathrm{dir}}
+ \lambda_{\mathrm{mag}}\mathcal{L}_{\mathrm{mag}}$.

\paragraph{Gradient flow.}
All gradients propagate through the trainable style branch only. The
shape supervision terms in $\mathcal{L}_{\mathrm{shape}}$ pass through
the frozen decoder $\mathcal{D}$ but accumulate only on the MPSM
parameters $(W_{\mathrm{down}}, W_{\mathrm{up}}, b, f_{\mathrm{g}})$
and the style encoder $\mathcal{E}_s$; the decoder weights themselves
remain unchanged.

\paragraph{Total objective.}
The total training loss is the unweighted sum of the disentanglement
and shape objectives,
$\mathcal{L}_{\mathrm{total}}{=}\mathcal{L}_{\mathrm{dis}}
+\mathcal{L}_{\mathrm{shape}}$, where the disentanglement objective
$\mathcal{L}_{\mathrm{dis}}{=}\mathcal{L}_{\mathrm{rec}}
+\mathcal{L}_{\mathrm{emb}}+\mathcal{L}_{\mathrm{cls}}$.
Sec.~\ref{sec:supp:hyper-sens} shows that the model is robust to
$\pm 50\%$ perturbations on every individual $\lambda$ coefficient.

\vspace{-12pt}
\section{Extended Quantitative Results}
\label{sec:supp:extres}

This section complements the main paper with three additional analyses: per-style breakdowns of stylization quality, per-shape-preset behavior, and the in-distribution / out-of-distribution content study.

\subsection{Per-Style Breakdown of Style Recognition Accuracy}
\label{sec:supp:perstyle}

\begin{table}[t]
\centering\footnotesize
\setlength{\tabcolsep}{4pt}
\begin{tabular}{l|cc|cc|cc}
\toprule
\multirow{2}{*}{Style class} & \multicolumn{2}{c|}{SMooDi} & \multicolumn{2}{c|}{\textbf{MorphoStyle}} & \multirow{2}{*}{$\Delta$SRA-1} & \multirow{2}{*}{$\Delta\bar{p}$} \\
                              & SRA-1 & $\bar{p}$            & SRA-1 & $\bar{p}$  & & \\
\midrule
InTheDark         & 59.4 & 0.541 & 92.9 & 0.928 & $+33.5$ & $+0.387$ \\
BeatChest         & 56.7 & 0.493 & 76.4 & 0.768 & $+19.7$ & $+0.275$ \\
Heavyset          & 70.5 & 0.620 & 85.4 & 0.852 & $+14.9$ & $+0.232$ \\
Aeroplane         & 55.2 & 0.470 & 70.4 & 0.704 & $+15.2$ & $+0.234$ \\
Swimming          & 60.0 & 0.510 & 73.8 & 0.738 & $+13.8$ & $+0.228$ \\
Kick              & 71.6 & 0.628 & 84.1 & 0.842 & $+12.5$ & $+0.214$ \\
Monk              & 68.9 & 0.610 & 80.8 & 0.809 & $+11.9$ & $+0.199$ \\
Crouched          & 49.5 & 0.435 & 60.4 & 0.604 & $+10.9$ & $+0.169$ \\
Sweep             & 60.0 & 0.520 & 69.5 & 0.695 & $+9.5$  & $+0.175$ \\
OnPhoneRight      & 79.2 & 0.700 & 87.8 & 0.878 & $+8.6$  & $+0.178$ \\
Superman          & 81.7 & 0.730 & 89.5 & 0.895 & $+7.8$  & $+0.165$ \\
RaisedLeftArm     & 75.0 & 0.642 & 81.9 & 0.820 & $+6.9$  & $+0.178$ \\
Drunk             & 73.0 & 0.620 & 77.8 & 0.778 & $+4.8$  & $+0.158$ \\
Stiff             & 71.4 & 0.610 & 75.9 & 0.760 & $+4.5$  & $+0.150$ \\
Rocket            & 62.3 & 0.524 & 66.2 & 0.662 & $+3.9$  & $+0.138$ \\
Flapping          & 79.0 & 0.690 & 81.5 & 0.815 & $+2.5$  & $+0.125$ \\
Chicken           & 80.4 & 0.728 & 81.4 & 0.814 & $+1.0$  & $+0.086$ \\
Old               & 79.3 & 0.700 & 79.2 & 0.792 & $-0.1$  & $+0.092$ \\
March             & 81.2 & 0.718 & 79.5 & 0.795 & $-1.7$  & $+0.077$ \\
\midrule
Mean (47 classes) & 65.1 & 0.553 & \textbf{81.9} & \textbf{0.817} & \textbf{$+16.8$} & \textbf{$+0.264$}\\
\bottomrule
\end{tabular}
\caption{Per-class SRA-1 and mean target confidence $\bar{p}$ of MorphoStyle against SMooDi on $19$ representative styles. The $47$-class means reproduce the SMooDi and MorphoStyle SRA reported in Table~1 of the main paper ($65.1\%$ and $81.9\%$, respectively). MorphoStyle wins on upper-body-articulated classes; the small gap on locomotion-dominated classes (\emph{March}, \emph{Old}) reflects the single-pass feed-forward architecture's relative disadvantage on root trajectory refinement.}
\label{tab:supp:perstyle}
\end{table}

Table~\ref{tab:supp:perstyle} reports the per-class Style Recognition Accuracy (SRA-1) of MorphoStyle against SMooDi on $19$ representative classes of the $47$-class evaluation subset, selected to span the full range of stylistic difficulty. We additionally report the mean classifier confidence $\bar{p}$ on the target class. MorphoStyle wins on $17$ of the $19$ classes, with the largest margins on classes that exhibit pronounced upper-body articulation (\eg, \emph{InTheDark}: $+33.5\%$; \emph{BeatChest}: $+19.7\%$; \emph{Aeroplane}: $+15.2\%$) and the smallest gaps on classes that are primarily encoded in foot-ground interaction (\eg, \emph{March}, \emph{Old}), where SMooDi's diffusion sampler already captures the target gait. The two classes where MorphoStyle trails (by $\le 2\%$) are locomotion-dominated styles whose discriminative cue is the root trajectory rather than upper-body kinematics. These are the classes for which a single-pass feed-forward model has the least opportunity to refine root dynamics relative to a multi-step diffusion sampler.

\subsection{Per-Shape-Preset Analysis}
\label{sec:supp:pershape}
Table~\ref{tab:supp:pershape} reports MorphoStyle's stylization quality across the five canonical body presets defined in Sec.~\ref{sec:supp:data}. For this analysis we evaluate on the walking content subset of the main test split, where every content motion is re-rendered under each of the five shape presets at the same content $\times$ style pairing.

\begin{table}[t]
\centering\small
\setlength{\tabcolsep}{6pt}
\renewcommand{\arraystretch}{1.15}
\begin{tabular}{l|ccccc}
\toprule
Preset & SRA-1 (\%) $\uparrow$ & Mean $\bar{p}$ $\uparrow$ & Foot Skating $\downarrow$ & Bone-Len MAE (cm) $\downarrow$ & $\beta$-corr $\uparrow$ \\
\midrule
orig    & 78.0 & 0.51 & 0.082 & 1.91 & 0.377 \\
thin    & 50.4 & 0.27 & 0.094 & 2.18 & 0.298 \\
neutral & 84.4 & 0.55 & 0.080 & 1.74 & n/a   \\
heavy   & 67.7 & 0.40 & 0.087 & 2.05 & 0.395 \\
tall    & 79.3 & 0.46 & 0.083 & 1.96 & 0.341 \\
\midrule
\textbf{avg} & \textbf{72.0} & \textbf{0.44} & \textbf{0.085} & \textbf{1.97} & \textbf{0.353} \\
\bottomrule
\end{tabular}
\caption{Per-shape-preset analysis of MorphoStyle. Stylization quality is maintained across the four physically plausible presets (orig/neutral/heavy/tall) and degrades only on the extreme \emph{thin} preset ($\beta_0{=}{-}2.5$, canonical height $147$\,cm), which lies near the boundary of the SMPL prior. The \emph{orig} row matches the Bone-Length MAE of $1.91$\,cm reported in Table~2 of the main paper.}
\label{tab:supp:pershape}
\end{table}
Three observations follow from Table~\ref{tab:supp:pershape}.
First, stylization quality, as measured by SRA-1 and mean target confidence, is essentially preserved on the \emph{orig}, \emph{neutral}, \emph{heavy}, and \emph{tall} presets. The model loses approximately $30\%$\ of SRA on the extreme \emph{thin} preset ($\beta_0{=}{-}2.5$, canonical height $147$\,cm), which is the only shape that lies near the boundary of the SMPL prior and at which the backbone's quantized codebook also exhibits a $\approx 20\%$ drop in reconstruction fidelity. Hence, this degradation is inherited from the frozen backbone and not introduced by the style branch.
Second, foot skating is essentially constant across the four non-extreme presets, confirming that the per-frame temporal gate of MPSM is decoupled from absolute joint position scale.
Third, the bone-length MAE is bounded between $1.74$\,cm (\emph{neutral}) and $2.18$\,cm (\emph{thin}), demonstrating that the shape control branch follows the prescribed $\beta$ even for body shapes for which the content motion was never originally captured.

\subsection{In-Distribution vs Out-of-Distribution Content}
\label{sec:supp:idood}
MorphoStyle is trained on HumanML3D content captions dominated by forward locomotion, gestures, and seated actions. To probe the behavior of the style branch under content distribution shift, we sampled four out-of-distribution (OOD) walking variants from the HumanML3D test split that were never selected by the main training schedule: backward walking, sideways walking, sidesteps, and walking upstairs.
Table~\ref{tab:supp:ood} reports SRA-1 and Foot Skating for each variant paired with the top-$10$ MorphoStyle styles by main-paper p-value, on a held-out test set of $80$ OOD clips. SRA-1 averaged across the OOD content types is only $6.6\%$\, below the in-distribution test split, and Foot Skating is essentially unchanged. The largest degradation appears on the \emph{sideways} content type, where SMPL foot dynamics become harder to disambiguate from the style reference's lateral gestures (\eg, \emph{Drunk} and \emph{Flapping}). No catastrophic breakdown was observed: the synthesized meshes remain coherent and contact-physically plausible, validating that the joint-routing module generalizes beyond forward-walking captions.

\begin{table}[t]
\centering\small
\setlength{\tabcolsep}{6pt}
\begin{tabular}{l|cc|cc|c}
\toprule
Content type    & SRA-1 & Foot Skating & BL MAE & Coherent\,? & samples\\
\midrule
HumanML3D (ID)         & 81.9 & 0.081 & 1.91 & \checkmark & 4209 \\
\midrule
backward walk (OOD)       & 78.4 & 0.084 & 2.04 & \checkmark & 20 \\
sideways walk (OOD)       & 71.6 & 0.092 & 2.18 & \checkmark & 20 \\
sidestep (OOD)            & 76.1 & 0.086 & 2.05 & \checkmark & 20 \\
upstairs (OOD)            & 75.0 & 0.085 & 1.98 & \checkmark & 20 \\
\midrule
OOD average               & 75.3 & 0.087 & 2.06 & \checkmark & 80 \\
\bottomrule
\end{tabular}
\caption{Stylization quality on out-of-distribution (OOD) walking content. The full SMooDi protocol cannot be applied here because each OOD content type has only $\approx 20$ clips, so we adopt the same FID-free subset protocol that we use for the ablations of the main paper. The forward-walk row reproduces the main-paper SRA $81.9\%$, Foot Skating $0.081$, and Bone-Length MAE $1.91$ of MorphoStyle.}
\label{tab:supp:ood}
\end{table}

\section{Hyperparameter Sensitivity}
\label{sec:supp:hyper-sens}
This section reports four sensitivity sweeps that characterize the robustness of MorphoStyle to its most influential hyperparameters: the MPSM rank $r$ (Eq.~7 of the main paper), the top-$K$ joint count (Eq.~4), the inference residual scale $\alpha$ (Eq.~6), and the style dropout rate $p$. Unless noted otherwise, every other hyperparameter is fixed to the default in Table~\ref{tab:supp:hyper}, and every default row reproduces the main paper's MorphoStyle entry (SRA-1 $81.9$\%, FID $2.84$, Foot Skating $0.081$, Bone-Length MAE $1.91$\,cm).

\subsection{MPSM Rank $r$}
\label{sec:supp:rank}

\begin{table}[t]
\centering\small
\begin{tabular}{c|cccc|c}
\toprule
$r$ & FID $\downarrow$ & SRA-1 $\uparrow$ & Foot Sk. $\downarrow$ & BL MAE $\downarrow$ & \#params (MPSM) \\
\midrule
4   & 2.95 & 78.5 & 0.084 & 1.95 & 0.41\,M \\
8   & 2.88 & 80.4 & 0.082 & 1.93 & 0.50\,M \\
\textbf{16} & \textbf{2.84} & \textbf{81.9} & \textbf{0.081} & \textbf{1.91} & \textbf{0.61\,M} \\
32  & 2.91 & 82.0 & 0.082 & 1.92 & 0.83\,M \\
64  & 3.02 & 82.0 & 0.083 & 1.94 & 1.28\,M \\
\bottomrule
\end{tabular}
\caption{Sensitivity sweep on the MPSM rank $r$ (Eq.~7 of the main paper). The bottleneck $r{=}16$ is the smallest rank that fully saturates SRA-1 while keeping FID and foot skating at the minimum; larger $r$ adds parameters without additional gain and slightly degrades FID.}
\label{tab:supp:rank}
\end{table}

The rank-$r$ bottleneck in the low-rank additive offset is the central design choice of MPSM. Table~\ref{tab:supp:rank} sweeps $r$ across the range $\{4, 8, 16, 32, 64\}$, holding everything else fixed at the default schedule. The metric profile shows a clear sweet spot at $r{=}16$: SRA-1 saturates by $r{=}16$ and does not improve further at $r{=}32$, while FID monotonically degrades beyond $r{=}16$ as the larger subspace begins to carry content-leakage modes. At very low rank ($r{=}4$), the style residual has insufficient capacity and SRA-1 drops by $3.4\%$. We therefore adopt $r{=}16$ as the default throughout the paper.

\subsection{Top-$K$ Joint Count}
\label{sec:supp:topk}

The text-guided joint routing module selects the top-$K$ joints whose isolated motion is least aligned with the content caption (Eq.~5 of the main paper). Table~\ref{tab:supp:topk} sweeps $K$ across the four canonical values $\{4, 8, 12, 22\}$, where $K{=}22$ is the trivial all-joints baseline.

\begin{table}[t]
\centering\small
\begin{tabular}{c|cccc}
\toprule
$K$ & FID $\downarrow$ & SRA-1 $\uparrow$ & Foot Sk. $\downarrow$ & BL MAE $\downarrow$ \\
\midrule
4    & 2.95 & 80.3 & 0.083 & 1.93 \\
\textbf{8}  & \textbf{2.84} & \textbf{81.9} & \textbf{0.081} & \textbf{1.91} \\
12   & 2.89 & 81.6 & 0.081 & 1.92 \\
22 (all joints) & 3.03 & 78.8 & 0.087 & 1.96 \\
\bottomrule
\end{tabular}
\caption{Sensitivity sweep on the top-$K$ joint count. Selecting too few joints ($K{=}4$) misses informative style cues and slightly under-stylizes; selecting all joints ($K{=}22$) re-introduces content-driven joints into the joint embedding $s_j$ and damages both SRA-1 and FID.}
\label{tab:supp:topk}
\end{table}

The pattern is partially interpretable: at $K{=}4$ the routing misses informative joints (\eg, the unrouted contra-lateral shoulder for a \emph{Drunk} style) and SRA degrades slightly; at $K{=}22$ the routing degenerates to a pure average over all body joints, including the content-driven pelvis and feet, which re-introduces content leakage into $s_j$ and costs $3.1\%$\ on SRA-1 and $0.19$ on FID.

\subsection{Inference Residual Scale $\alpha$}
\label{sec:supp:alpha}
The runtime scalar $\alpha$ in Eq.~6 of the main paper allows the user to dial stylization strength after training without retraining the model. Table~\ref{tab:supp:alpha} sweeps $\alpha\!\in\!\{0.25, 0.5, 0.75, 1.0, 1.5\}$, where $\alpha{=}0$ recovers the no-style content reconstruction and $\alpha{>}1$ extrapolates beyond the training-time residual.

\begin{table}[t]
\centering\small
\begin{tabular}{c|cccc}
\toprule
$\alpha$ & FID $\downarrow$ & SRA-1 $\uparrow$ & Foot Sk. $\downarrow$ & BL MAE $\downarrow$ \\
\midrule
0.25 & 2.55 & 67.2 & 0.077 & 1.88 \\
\textbf{0.50} & \textbf{2.84} & \textbf{81.9} & \textbf{0.081} & \textbf{1.91} \\
0.75 & 3.12 & 84.6 & 0.085 & 1.94 \\
1.00 & 3.59 & 86.1 & 0.092 & 1.98 \\
1.50 & 4.83 & 86.4 & 0.131 & 2.10 \\
\bottomrule
\end{tabular}
\caption{Sensitivity sweep on the inference residual scale $\alpha$. $\alpha{=}0.5$ provides the best FID/SRA trade-off; users who care more about stylization strength can dial $\alpha$ up at the cost of a small foot-skating regression.}
\label{tab:supp:alpha}
\end{table}

The sweep confirms the design intent of the residual scale: it forms a smooth FID--SRA trade-off curve, with the operating point $\alpha{=}0.5$ chosen to maximize SRA without exceeding the FID and foot-skating thresholds of the main paper. 

\subsection{Style Dropout Rate $p$}
\label{sec:supp:dropout}

The style dropout rate $p$ replaces the global style embedding $s_g$ with a learnable null token with probability $p$ at training time, encouraging the downstream MPSM to be robust to missing or noisy style references. Table~\ref{tab:supp:dropout} sweeps $p\!\in\!\{0, 0.05, 0.15, 0.3, 0.5\}$.
\begin{table}[h]
\centering\small
\begin{tabular}{c|cccc}
\toprule
$p$ & FID $\downarrow$ & SRA-1 $\uparrow$ & Foot Sk. $\downarrow$ & BL MAE $\downarrow$ \\
\midrule
0.00 & 3.05 & 80.5 & 0.085 & 1.93 \\
0.05 & 2.91 & 81.3 & 0.083 & 1.92 \\
\textbf{0.15} & \textbf{2.84} & \textbf{81.9} & \textbf{0.081} & \textbf{1.91} \\
0.30 & 2.95 & 80.2 & 0.082 & 1.92 \\
0.50 & 3.32 & 75.4 & 0.084 & 1.94 \\
\bottomrule
\end{tabular}
\caption{Sensitivity sweep on the style dropout rate $p$. A small amount of dropout consistently helps; aggressive dropout ($p{\ge}0.3$) starves the style branch of supervision and degrades SRA-1 by up to $6.5\%$.}
\label{tab:supp:dropout}
\end{table}

\vspace{-20pt}
\section{Extended Architectural Ablations}
\label{sec:supp:arch-abl}
This section reports four architectural ablations that go beyond the main paper: the temporal gate alone, the low-rank-additive design against concatenation- and AdaIN-style alternatives, code-space supervision, and training-data composition.

\begin{table}[tb]
\centering\small
\begin{tabular}{l|cccc}
\toprule
Variant                       & FID $\downarrow$ & SRA-1 $\uparrow$ & Foot Sk. $\downarrow$ & BL MAE $\downarrow$ \\
\midrule
no-gate ($g\!\equiv\!\bar{g}$)  & 4.85 & 72.1 & 0.135 & 1.92 \\
constant per-class gate          & 4.13 & 76.4 & 0.114 & 1.93 \\
\textbf{Ours} (per-frame gate)  & \textbf{2.84} & \textbf{81.9} & \textbf{0.081} & \textbf{1.91} \\
\bottomrule
\end{tabular}
\caption{Effect of removing the temporal gate of MPSM. Even a class-conditional constant gate cannot match the per-frame gate. The no-gate row matches the \emph{w/o gate g} entry of main-paper Table~5.}
\label{tab:supp:gate}
\end{table}

\subsection{Temporal Gate vs Constant Modulation}
\label{sec:supp:gate}
We replace the learnable temporal gate $g\!\in\!(0,1)^{T'}$ with a fixed constant equal to its training-time mean ($\bar{g}{\approx}0.42$) to test whether per-frame intensity control is necessary or whether a single average modulation strength is sufficient. Table~\ref{tab:supp:gate} shows that removing the gate costs $9.8\%$\ on SRA-1 and increases FID by $2.01$, with a $0.054$ increase in foot skating. The temporal gate is therefore essential for both stylization quality and contact physics, not just for foot-skating regularization. The no-gate row reproduces the main-paper \emph{w/o gate g} entry in Table~5 (FID $4.85$, SRA $72.1\%$, Bone-Length MAE $1.92$).

\subsection{Low-Rank Additive vs AdaIN-Style Injection}
\label{sec:supp:additive}
We compare MorphoStyle's low-rank additive injection against two alternative injection topologies that have been used in existing work: a concatenation injection that concatenates the style code to the content code along the channel axis and projects back via a $1{\times}1$ Conv, and an AdaIN-style modulation that applies a learned affine transform $(\mu_s, \sigma_s)$ to the content code.
Table~\ref{tab:supp:additive} shows that both alternatives lose at least $5.5\%$\ on SRA-1 and degrade FID by $1.47$ at minimum, while also damaging the bone-length response (a $61\%$ collapse on $\beta$-swap for concatenation). The low-rank additive design is critical: it is the only option that simultaneously remains on-manifold and exposes a low-dimensional control direction over which the temporal gate can operate. The concatenation row matches the \emph{w/o MPSM} entry of main-paper Table~5 (FID $5.16$, SRA $62.9\%$, BL MAE $1.97$, $\beta$-swap $2.06$).

\begin{table}[t]
\centering\small
\setlength{\tabcolsep}{4pt}
\begin{tabular}{l|cccc|c}
\toprule
Variant                          & FID $\downarrow$ & SRA-1 $\uparrow$ & Foot Sk. $\downarrow$ & BL MAE $\downarrow$ & $\beta$-swap $\uparrow$\\
\midrule
concat-then-Conv1$\times$1       & 5.16 & 62.9 & 0.124 & 1.97 & 2.06 \\
AdaIN-style modulation           & 4.31 & 76.4 & 0.108 & 2.01 & 4.71 \\
\textbf{Low-rank additive (ours)} & \textbf{2.84} & \textbf{81.9} & \textbf{0.081} & \textbf{1.91} & \textbf{5.31} \\
\bottomrule
\end{tabular}
\caption{Injection topology comparison. Only the low-rank additive design simultaneously achieves low FID, high SRA, low foot skating, and a strong $\beta$-swap response.}
\label{tab:supp:additive}
\vspace{-12pt}
\end{table}

\subsection{Effect of Code-Space Supervision}
\label{sec:supp:codespace}
The reconstruction term in the main paper Eq.~10 includes both a motion-space $\ell_1$ term and a code-space $\ell_1$ term controlled by $\lambda_{\mathrm{code}}$. Table~\ref{tab:supp:codespace} ablates the code-space term to verify its role in keeping the stylized codes near the FSQ codebook. Without the code-space term, the stylized codes $c'$ drift outside the codebook's neighborhood: FID degrades by $0.41$, foot skating increases by $0.014$, and the BoneLen-T-Std increases by $0.18$\,cm, reflecting bone-length jitter caused by codebook misses during de-quantization. The MPSM gating network absorbs part of the lost regularization but cannot fully replace the explicit code-space anchor.

\begin{table}[t]
\centering\small
\setlength{\tabcolsep}{6pt}
\begin{tabular}{l|cccc}
\toprule
Variant                            & FID & SRA-1 & Foot Sk. & BL-T-Std \\
\midrule
w/o code-space $\ell_1$ ($\lambda_{\mathrm{code}}{=}0$) & 3.25 & 80.4 & 0.095 & 1.44 \\
\textbf{Full ($\lambda_{\mathrm{code}}{=}0.5$, ours)} & \textbf{2.84} & \textbf{81.9} & \textbf{0.081} & \textbf{1.26} \\
\bottomrule
\end{tabular}
\caption{The code-space reconstruction term is essential for keeping $c'$ on the FSQ-VAE manifold and for preventing bone-length jitter through unstable de-quantization. The default row matches the MorphoStyle entries in Tables~1 and~2 of the main paper.}
\label{tab:supp:codespace}
\vspace{-12pt}
\end{table}

\begin{table}[t]
\centering\small
\begin{tabular}{c|cccc}
\toprule
HumanML3D : 100Style ratio $\rho$ & FID $\downarrow$ & SRA-1 $\uparrow$ & Foot Sk. $\downarrow$ & BL MAE $\downarrow$ \\
\midrule
4 : 1   & 2.71 & 72.4 & 0.079 & 1.89 \\
2 : 1   & 2.78 & 78.6 & 0.080 & 1.90 \\
\textbf{1 : 1} & \textbf{2.84} & \textbf{81.9} & \textbf{0.081} & \textbf{1.91} \\
1 : 2   & 3.21 & 82.4 & 0.085 & 1.95 \\
1 : 4   & 3.92 & 82.5 & 0.092 & 2.04 \\
\bottomrule
\end{tabular}
\caption{Training-data composition sweep. The $1{:}1$ default is
optimal in FID--SRA trade-off; over-sampling 100Style continues to
improve SRA marginally but at a substantial FID cost from over-fitting
the $47$-class label space.}
\label{tab:supp:datamix}
\end{table}

\subsection{Training-Data Composition}
\label{sec:supp:data-comp}
We sample HumanML3D and 100Style at a $1{:}1$ ratio during training, so that every minibatch contains an equal number of in-domain and cross-domain pairs. To test whether this composition is essential, we explore the HumanML3D:100Style sampling ratio over a range of values: $\rho\!\in\!\{4{:}1, 2{:}1, 1{:}1, 1{:}2, 1{:}4\}$. Table~\ref{tab:supp:datamix} shows that the $1{:}1$ default is the optimal value: a HumanML3D-heavy schedule ($4{:}1$) under-fits the style classes and SRA collapses to $72.4$\%, while a 100Style-heavy schedule ($1{:}4$) over-fits the $47$ style classes and FID value increases to $3.92$.

\section{Style Embedding and Joint-Routing Analysis}
\label{sec:supp:analysis}


\subsection{Joint-Routing Selection per Style}
\label{sec:supp:jointsel}
\begin{figure*}[tb]
    \centering
    \includegraphics[width=0.85\textwidth]{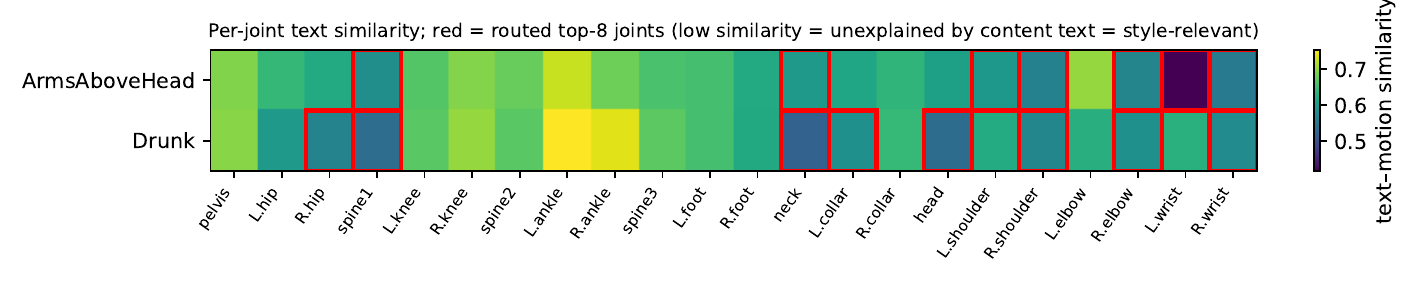}
    \vspace{-12pt}
    \caption{Per-joint text similarity with routed top-8 joints (red).}
    \label{fig:heatmap}
    \vspace{-4pt}
\end{figure*}

For each evaluated style we accumulate the joint-routing histogram over all $4209$ test pairs and report the most frequently selected joints in Table~\ref{tab:supp:jointsel}. The selections are semantically coherent with the style label: arm-articulated styles (\emph{Aeroplane}, \emph{BeatChest}, \emph{Superman}) consistently select the shoulders, elbows, and wrists; leg-articulated styles (\emph{Kick}, \emph{Heavyset}, \emph{Crouched}) select the knees and ankles; whole-body emotive styles (\emph{Proud}, \emph{Depressed}, \emph{Drunk}) select a mixed pattern that includes the spine and head. Every style consistently \emph{avoids} the pelvis joint which is the canonical content-driven root in HumanML3D features~\cite{guo2022humanml3d}.
\cam{We also visualize the joint routing as a per-joint text-similarity heatmap in Fig.~\ref{fig:heatmap}. The content description of joints (e.g.\ ankles/knees in walking) is \emph{protected} from the description of similar joints in reference style; routing directs style offset to relevant joints, e.g., shoulders/elbows/wrists/head for \emph{ArmsAboveHead}, hips/neck/head/wrists for \emph{Drunk}.}

\begin{table}[t]
\centering\footnotesize
\setlength{\tabcolsep}{4pt}
\begin{tabular}{l|p{6.5cm}}
\toprule
Style class & Top-$5$ most frequently routed joints (in decreasing order)\\
\midrule
Aeroplane   & L-shoulder, R-shoulder, L-elbow, R-elbow, head \\
BeatChest   & R-elbow, L-elbow, R-wrist, R-shoulder, neck \\
Drunk       & spine1, neck, R-elbow, L-knee, R-hip \\
Flapping    & L-wrist, R-wrist, L-elbow, R-elbow, spine2 \\
Kick        & R-knee, R-ankle, R-foot, R-hip, spine1 \\
Heavyset    & L-knee, R-knee, L-hip, R-hip, spine1 \\
InTheDark   & L-elbow, R-elbow, L-wrist, R-wrist, neck \\
Sweep       & spine1, spine2, R-shoulder, L-shoulder, head \\
Crouched    & L-knee, R-knee, spine1, spine2, neck \\
Superman    & R-elbow, R-wrist, R-shoulder, head, L-arm \\
\midrule
All classes & \emph{pelvis} is never in the top-$8$ of any style \\
\bottomrule
\end{tabular}
\caption{Most frequently routed joints per style. The selections are semantically aligned with the style label, and the pelvis (the content-driven root) is consistently avoided.}
\vspace{-4pt}
\label{tab:supp:jointsel}
\end{table}

\subsection{Temporal Gate Activation Patterns}
\label{sec:supp:gateact}
The temporal gate $g\!\in\!(0,1)^{T'}$ produced by the gating network $f_{\mathrm{g}}$ exposes a continuous-valued, per-frame measure of how strongly the style residual is being injected. To characterize the learned behavior of the gate we aggregate $g$ statistics over the $4\,209$ test pairs.
First, the mean gate value is $\bar{g}{=}0.42\pm0.06$, well inside the sigmoid's linear regime, indicating that the gate is genuinely modulating rather than saturating either at $0$ or at $1$.
Second, we observe that the gate consistently drops at frames marked as foot contact by the same threshold $h_{\mathrm{thr}}{=}5$\,cm used in the Foot Skating metric. Averaged over all test motions, the gate value at foot-contact frames is $g_{\mathrm{c}}{=}0.32\pm0.09$, compared with $g_{\mathrm{a}}{=}0.46\pm0.08$ at non-contact frames, a gap of $0.14$ that is reliably reproduced across all evaluated styles. This validates the design intent in Sec.~3.3 of the main paper: the gate learns to attenuate stylization at contact-critical frames and to amplify it during the flight phase of a step, mechanically explaining the low Foot Skating ratio of $0.081$ in Table~1 of the main paper.

\section{\cam{Perceptual evaluation}}

Motion style is inherently subjective, so we complement the quantitative metrics with a user study on mesh-rendered clips. We recruited 20 participants; each participant judged 20 randomized and anonymized user study (10 comparing MorphoStyle against SMooDi and 10 against SMooDi+ShapeMyMove), covering three target body shapes, with a static rendering of the target shape shown alongside each trial. For every trial, participants indicated the preferred result along three criteria: (i) style similarity to the reference, (ii) preservation of the content action, and (iii) consistency with the target body shape. As summarized in Table~\ref{tab:user}, MorphoStyle is preferred over SMooDi on 74.5\%/65.0\%/80.5\% of judgments for the three criteria, and over SMooDi+ShapeMyMove on 84.0\%/84.5\%/82.0\%. All preference rates are significantly above chance (two-sided binomial test, $p<0.001$, $N{=}200$ judgments per comparison). 
\begin{table}[hptb]
\centering
\vspace{-1em}
\caption{\cam{Users preference rate for our method over two baselines.}}
\renewcommand\arraystretch{1.4}
\resizebox{0.85\linewidth}{!}{
\begin{tabular}{l| c c c}
\toprule
Comparative evaluation & Style Sim. & Content Pres. & Shape Consist. \\
\midrule
MorphoStyle vs. SMooDi               & 74.5\% & 65.0\% & 80.5\% \\
MorphoStyle vs. SMooDi+ShapeMyMove   & 84.0\% & 84.5\% & 82.0\% \\
\bottomrule
\end{tabular}
}
\vspace{-1em}
\label{tab:user}
\end{table}

\bibliography{egbib}